\pdfoutput=1
\documentclass[10pt,twocolumn,letterpaper]{article}

\usepackage{iccv}
\usepackage{times}
\usepackage{epsfig}
\usepackage{graphicx}
\usepackage{amsmath}
\usepackage{amssymb}
\usepackage{color}
\usepackage{multirow}
\usepackage{cite}
\usepackage{float}
\usepackage{array}
\usepackage{booktabs}
\usepackage{siunitx}
\usepackage{longtable}
\usepackage{footnote}
\usepackage{tablefootnote}
\usepackage[caption=false]{subfig}
\usepackage[inline]{enumitem}
\usepackage[table,xcdraw]{xcolor}
\usepackage{makecell}
\usepackage[accsupp]{axessibility} 

\usepackage[pagebackref=true,breaklinks=true,letterpaper=true,colorlinks,bookmarks=false]{hyperref}

\colorlet{LightGray}{gray!10}
\colorlet{MediumGray}{gray!30}
\colorlet{DarkGray}{gray!50}

\iccvfinalcopy 

\def\iccvPaperID{9272} 

\ificcvfinal\fi

\begin{document}

\title{CrowdDriven: A New Challenging Dataset for Outdoor Visual Localization}

\author{
Ara Jafarzadeh$^1$
\quad
Manuel López Antequera$^2$
\quad
Pau Gargallo$^2$
\quad
Yubin Kuang$^2$
\quad
Carl Toft$^1$\\
Fredrik Kahl$^1$
\quad
Torsten Sattler$^3$\\
{
$^1$Chalmers University of Technology \hspace{16pt}
$^2$Facebook \hspace{16pt}
$^3$Czech Technical University in Prague
}
}

\maketitle
\ificcvfinal\thispagestyle{empty}\fi

\vspace{-12pt}
\begin{abstract}
\vspace{-8pt}
   Visual localization is the problem of estimating the position and orientation from which a given image (or a sequence of images) is taken in a known scene. It is an important part of a wide range of computer vision and robotics applications, from self-driving cars to augmented/virtual reality systems. Visual localization techniques should work reliably and robustly under a wide range of conditions, including seasonal, weather, illumination and man-made changes. Recent benchmarking efforts model this by providing images under different conditions, and the community has made rapid progress on these datasets since their inception.
   However, they are limited to a few geographical regions and
   often recorded with a single device. 
   We propose a new benchmark for visual localization in outdoor scenes, using crowd-sourced data to cover a wide range of geographical regions and camera devices with a focus on the failure cases of current algorithms.
   Experiments with state-of-the-art localization approaches show that our dataset is very challenging, with all evaluated methods failing on its hardest parts.
   As part of the dataset release, we provide the tooling used to generate it, enabling efficient and effective 2D correspondence annotation to obtain reference poses.

\end{abstract}

\vspace{-12pt}
\section{Introduction}

Visual localization is the problem of estimating the position and orientation from which an image was taken, \ie, its camera pose, with respect to the scene. 
Visual localization is a vital part of many computer vision and robotics applications such as self-driving cars, service robots such as gardening robots, and augmented/mixed/virtual reality. 

\begin{figure*}[tb]
    \centering
    \includegraphics[width=\linewidth]{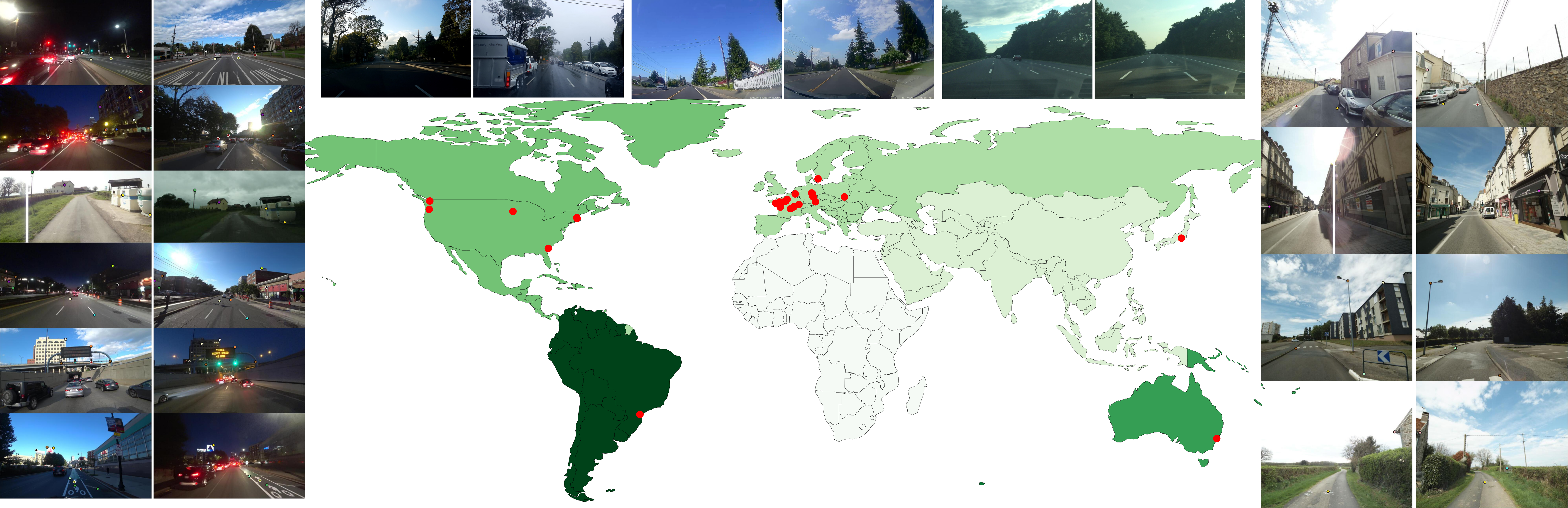}
    \caption{Easy (top), Medium (left) and Hard (right) subsets of our CrowdDriven dataset. Our dataset includes reference poses derived from hand-annotated control points, enabling the benchmarking of localization algorithms in new scenarios such as extreme viewpoint (180$^\circ$) changes. CrowdDriven includes sequences from over the globe captured with a diversity of cameras.}
    \label{fig:geodist}
\end{figure*}

Most visual localization methods rely heavily on local descriptors for pose estimation, and finding 2D-3D matches between images is a fundamental part of it. However, the trade-off between the local descriptors' discriminative power and their invariance limits their performance under changing conditions. On the other hand, in practice, changes in the scene are unavoidable, and there is a need for visual localization methods to be robust to them. 
Traditionally, ground truth poses for localization have been obtained via Structure-from-Motion (SfM)~\cite{Schoenberger2016CVPR,Snavely-IJCV08,Heinly2015CVPR}. 
Yet, SfM 
itself relies on 
local descriptors and matching. This makes it extremely difficult to generate localization benchmarks where feature matching using traditional feature descriptors does not work, \eg, in the presence of day/night and strong viewpoint changes. Benchmark datasets for localization under changing conditions such as Aachen, CMU Seasons, and Robotcar \cite{Sattler2018CVPR} rely on manual annotations to be able to provide ground truth poses. 
While these datasets provide interesting challenges
, they have mostly been captured in
controlled conditions.
However, there is little control over the capture in many applications such as Autonomous Driving, collaborative AR/MR, or crowd-sourced mapping.

In this paper, we have constructed a dataset revisiting the common challenges that could be seen in different environments. We first actively mined a crowd-sourced database for image sequences where classic SfM approaches fail. To generate reliable poses, we have relied on human annotations and have created 40 sets of image sequences with diverse visual changes. The dataset and tools used for its creation are available at \href{https://www.mapillary.com/datasets}{mapillary.com}. 
This paper makes the following contributions: 
(1) a workflow to mine and annotate challenging image sequences for benchmarking visual localization. 
Our approach explicitly takes pose uncertainty into account during the annotation process. 
(2) the CrowdDriven dataset, a geographically diverse and challenging dataset 
with reliable poses that covers various scenarios in illumination, weather, seasonal, and viewpoint changes. 
(3) experiments with state-of-the-art baselines showing CrowdDriven presents challenges that existing methods cannot handle.

\vspace{-6pt}
\section{Related Work}
\vspace{-3pt}

\noindent\textbf{Visual localization} methods aim to estimate the full camera pose and can be categorized based on their map representation:

(1)~\textbf{Image-based representations} encode each image as a feature vector~\cite{ge-eccv-2020,Arandjelovic2016CVPR,Chen2017ICRA,Torii2015CVPR,torii11,Sattler2016CVPR}. 
Using image retrieval and the known poses of the database images, the pose of a test image can either be approximated via the poses of the top-retrieved database images~\cite{Zamir10ECCV,Sattler2019CVPR} or computed precisely based on relative poses~\cite{Sattler2017CVPR,Zhang06TDPVT}.

(2)~\textbf{Local feature-based representations} rely on classical or learned local feature descriptors and multiple view geometry \cite{se2002mobile, li2010location,Sattler2011ICCV,Li2012ECCV,Sattler2012ECCV,Zeisl2015ICCV,Sattler2017PAMI,Svarm2017PAMI}. These approaches build a 3D model of the scene and then estimate the camera pose based on 2D-3D matches between local features in a test image and the 3D model~\cite{Kukelova2016CVPR,Haralick94IJCV,Fischler81CACM}. Such approaches, especially using learned features, constitute the current state-of-the-art in terms of visual localization in changing conditions~\cite{dusmanu2019d2,germain2019sparse,sarlin2019coarse,sarlin2020superglue,schonberger2018semantic} and we evaluate and analyze their performance on our benchmark.

(3) \textbf{Learning-based representations} typically represent the scene using  convolutional neural networks (CNNs). 
Pose regression techniques directly regress the camera pose for a given input image~\cite{Kendall2015ICCV,Kendall2017CVPR,Walch2017ICCV,Brahmbhatt2018CVPR}. These methods have been shown not to perform better than image retrieval methods~\cite{Sattler2019CVPR}. 
Learning-based approaches that do not learn the full localization pipeline but only the 2D-3D matching part~\cite{Shotton2013CVPR,Brachmann2016CVPR,Brachmann2017CVPR,Brachmann2020ARXIV,Cavallari2017CVPR,Brachmann2019ICCV,Cavallari20193DV,Sarlin2021CVPR,Dong2021CVPR} achieve state-of-the-art performance in small scenes~\cite{Brachmann2021ICCV,wald-eccv-2020}. 
While most of these approaches struggle to handle conditions not seen during training, \cite{Sarlin2021CVPR} generalizes well.

\noindent While localization approaches use metric maps, \textbf{visual place recognition} methods typically rely on topological scene representations~\cite{milford2012seqslam, vysotska2015lazy, naseer2018robust, doan2020visual, garg2019semantic,doan2019scalable, garg2021semantics}, and thus do not directly provide a camera pose estimate. 
Using image retrieval and related techniques, they identify the \textit{places} depicted in test images, where \textit{places} are defined as sets of database images. 
Place recognition can be used to guide visual localization by identifying which parts of a scene are visible in a test image~\cite{irschara2009structure,Sattler2012BMVC,sarlin2019coarse}. 
For an overview over place recognition techniques, we refer the reader to~\cite{lowry2016visual, garg2021your}.


\begin{table*}[tb]
\centering
\resizebox{\linewidth}{!}{
\begin{tabular}{|c|c|c|c|c|c|c|c|c|c|c|c|c|c|c|c|c|c|}
\toprule
\multirow{2}{*}{Dataset} & \multicolumn{5}{c|}{Scene Type}                                          & \multicolumn{2}{c|}{\#images} & \multicolumn{7}{c|}{condition changes}                                                                                      & \multirow{2}{*}{sequential} & \multirow{2}{*}{6DOF query poses} & \multirow{2}{*}{\# locations} \\ \cline{2-15}
    & urban              & suburban           & natural            & country road               & indoors   & reference     & query         & weather            & seasonal           & strong viewpoint   & day/night          & intrinsics         & snow               & rain               &        &                                   \\  \hline
Nordland\cite{sunderhaufwe}                 & \checkmark          & \checkmark          & \checkmark          &                    &           & 14k           & 16k           &                    & \checkmark          &                    &                    &                    &                    &                    &  \checkmark    &             &     1         \\
Pittsburgh \cite{torii2013visual}              & \checkmark          &                    &                    &                    &           & 254k              & 24k               &                    &                    &                    &                    &                    &                    &                    &   \checkmark   &              &    1                 \\
Tokyo 24/7\cite{Torii2015CVPR}               & \checkmark          &                    &                    &                    &           & 174k          & 1k            &                    & \checkmark          &                    & \checkmark          &     \checkmark            &                    &                    &        &                         & 1         \\
NCLT\cite{carlevaris2016university}& \checkmark          &                    &     \checkmark              &                    & \checkmark & \multicolumn{2}{c|}{3.8M}     &                    & \checkmark          &                    &                    &                    &      \checkmark         &                    &  \checkmark    &  \checkmark  &             1       \\
Extended CMU Seasons\cite{Sattler2018CVPR,Badino_IV11}         & \checkmark          & \checkmark          &    \checkmark                &     \checkmark         &           & 61k           & 57k           & \checkmark          & \checkmark          &                    &          &        \checkmark          &         \checkmark       &                    &    \checkmark  &     \checkmark    &    1             \\
RobotCar Seasons\cite{Sattler2018CVPR,Maddern2017IJRR}         & \checkmark          & \checkmark          &                    &                    &           & 20k           & 12k           & \checkmark          & \checkmark          &                    & \checkmark          &         \checkmark         &          \checkmark       &      \checkmark           &   \checkmark    &   \checkmark    &    1             \\
Aachen Day-Night\cite{Sattler2018CVPR,Sattler2012BMVC}                   & \checkmark          &                    &                    &                    &           & 3k            & 922           &                    &                    &                    & \checkmark          &     \checkmark            &                    &                    &        & \checkmark & 1                         \\
RIO10\cite{wald-eccv-2020}                   &           &                    &                    &                    &  \checkmark         & 53k            & 200k           &                    &                    &                    &           &                    &                    &                    &     \checkmark   &        \checkmark       &    1    \\
7-scenes\cite{7scenes}                   &           &                    &                    &                    &  \checkmark         & 26k & 17k      &                    &                    &                    &           &                    &                    &                    &   \checkmark      &        \checkmark       &    7    \\
12-scenes\cite{12scenes}                   &           &                    &                    &                    &  \checkmark         &  17k &     5.8k    &                    &                    &                    &           &                    &                    &                    &   \checkmark     &        \checkmark       &    12    \\
Cambridge\cite{Kendall2015ICCV}                   &   \checkmark        &                    &                    &                    &          &  8.4k & 4.8k          &                    &                    &                    &           &                    &                    &                    &   \checkmark     &        \checkmark       &    3    \\
Dubrovnik\cite{li2010location}               & \checkmark          &                    &                    &                    &           & 6k            & 0.8k          &                    &                    &                    &                    &                    &                    &                    &        & \checkmark        & 1                 \\
San Francisco\cite{chen2011city}            & \checkmark          &                    &                    &                    &           & 610k          & 0.4k          &                    &                    &                    &                    &                    &                    &                    &        & \checkmark          & 1               \\
Rome\cite{li2010location}& \checkmark          &                    &                    &                    &           & 15k           & 1k            &                    &                    &                    &                    &                    &                    &                    &        &                      & 1             \\
Vienna\cite{irschara2009structure}                & \checkmark          &                    &                    &                    &           & 1k            & 0.2k          &                    &                    &                    &                    &                    &                    &                    &        &                &  1                 \\
InLoc\cite{Taira2018CVPR,wijmans17rgbd}                &           &                    &                    &                    &     \checkmark       & 9.9k            & 0.3k          &                    &                    &                    &                    &                    &                    &                    &    \checkmark           &   \checkmark     &          5          \\
\midrule
\textbf{CrowdDriven}     & \textbf{\checkmark} & \textbf{\checkmark} & \textbf{\checkmark} & \textbf{\checkmark} & \textbf{} & \textbf{1.3k} & \textbf{1.7k} & \textbf{\checkmark} & \textbf{\checkmark} & \textbf{\checkmark} & \textbf{\checkmark} & \textbf{\checkmark} & \textbf{\checkmark} & \textbf{\checkmark} & \textbf{\checkmark}     & \textbf{\checkmark}   &   \textbf{26}          \\
\bottomrule
\end{tabular}
}
\caption{Comparing localization datasets: CrowdDriven is the most diverse in term of scene types and changes in viewing conditions.}
\label{tab:dataset_comparison}
\end{table*}

 Tab.~\ref{tab:dataset_comparison} provides an overview over \textbf{datasets commonly used to measure localization and place recognition performance under changing conditions}. 
Coarse-scale location information for place recognition datasets such as Nordland~\cite{sunderhaufwe}, Pittsburgh~\cite{torii2013visual}, Tokyo~24/7\cite{Torii2015CVPR}, and Mapillary Street-Level Sequences~\cite{warburg2020mapillary} can be obtained relatively easily via GPS measurements. 
In contrast, obtaining 6DOF 
poses for localization requires considerable manual effort as the classical approaches used to automatically obtain ground truth fail under challenging conditions.
RIO10 \cite{wald-eccv-2020} focuses on changes in indoor scenes.
Long-term outdoor localization datasets such as Aachen Day-Night~\cite{Sattler2012BMVC,Sattler2018CVPR}, CMU Seasons~\cite{Sattler2018CVPR,Badino_IV11}, and RobotCar Seasons~\cite{Sattler2018CVPR,Maddern2017IJRR} only cover a few geographical locations and using a small number of cameras. In contrast, even though our new benchmark dataset is not the largest, it has much more geographical diversity. 
While the other datasets were captured using only a few cameras, our image sequences are taken by a large number of different camera types and photographers.

Other localization datasets 
include the indoor 7 Scenes~\cite{Shotton2013CVPR}, 12 Scenes~\cite{Valentin2016}, and InLoc~\cite{Taira2018CVPR} datasets and the outdoor Dubrovnik~\cite{li2010location}, Rome~\cite{Frahm10ECCV}, Vienna~\cite{irschara2009structure}, San Francisco~\cite{Li2012ECCV,chen2011city}, and Cambridge Landmarks~\cite{Kendall2015ICCV} datasets. 
None of these datasets is designed to measure the impact of changing conditions on localization performance~\cite{Sattler2018CVPR}.

The focus of recent datasets from self-driving car companies such as Lyft\cite{lyft2019}, Waymo \cite{Waymo2019Open}, Aptiv (nuScenes) \cite{Caesar2019CoRR}, and  Baidu (Apolloscape) \cite{apolloscape} 
is benchmarking 2D/3D object detection based on 
multimodal sensor data in limited geographical areas, 
semantic segmentation, and depth estimation. An exception is Baidu's Apolloscape which provides a localization benchmark track in small urban areas.

Another shortcoming of the datasets mentioned above is that the changes in viewpoints are limited. The query images are typically taken from similar vantage points as the reference images. Our dataset contains image sequences that are, for instance, taken in opposite directions
with different cameras. 
This case commonly happens during multi-session or collaborative capture of man-made environments, where paths can be traversed in two directions.

Based on the shortcomings of current localization datasets, we generate a diverse dataset for long-term visual localization (1) containing different challenging scenarios such as day-night, seasonal, daylight illumination, strong viewpoint and man-made changes, (2) with reliable 6DOF camera poses based on human annotations, (3) using crowd-sourced data to imitate real driving situations, (4) with reliable camera calibration and geographical information.

\vspace{-4pt}
\section{CrowdDriven: Dataset Creation}
\label{sec:dataset}
\vspace{-4pt}

Historically, visual localization algorithms have been evaluated on crowd-sourced datasets collected from internet photo 
collection websites such as Flickr, \eg, the Dubrovnik~\cite{li2010location}, Rome~\cite{li2010location}, and Landmarks 1k~\cite{Li2012ECCV} datasets. 
This way to acquire datasets scales very well to the use of different camera types and geographical locations. 
However, since the images have been taken by handheld cameras around famous landmarks, these datasets are not suitable for measuring visual localization performance in an autonomous driving scenario, which is our main focus 
(even though a small part of our dataset was captured by pedestrians and  bicycles).   
In contrast, recently released datasets tailored for this task, \eg, the RobotCar Seasons~\cite{Sattler2018CVPR,Maddern2017IJRR} and the (extended) CMU Seasons~\cite{Sattler2018CVPR,Badino_IV11} datasets, contain only one or a few larger locations, use only a few cameras, and have been taken by experts. 
Our new dataset, CrowdDriven, is designed to cover a wide range of locations, visual conditions (\eg, seasonal, day-night changes or changes in facades, presence/absence of humans (man-made changes) ),
and camera types by tapping into a large database of crowd-sourced images.
\vspace{-3pt}
\subsection{Data Source} 
\vspace{-3pt}
To maximize the diversity and geographical coverage, we use 
\href{https://www.mapillary.com}{Mapillary},  

a collaborative street-level imagery platform that hosts more than 1 billion images collected by members of their community while driving or walking on public spaces and roads. It covers most countries with hundreds of 
camera models in varying times of day and weather conditions. 
Its data is thus well-suited to evaluate problems in conditions similar to those faced in self-driving scenarios, since most images are captured with consumer-grade devices such as smartphones, action cameras, and dashcams.
\vspace{-3pt}
\subsection{Sequence Selection}
\vspace{-3pt}

Images in Mapillary are grouped into ordered sequences based on the photographer ID and the capture time. Each image's GPS position and compass are available, and the database can be queried to find images near a particular position.
To generate a dataset that is challenging for current state-of-the-art localization algorithms, we attempt to find pairs of neighboring sequences to be localized with respect to each other.
We categorize each sequence pair into a preliminary difficulty level based on the success of traditional SfM algorithms on jointly reconstructing the pair. 

We begin by querying the Mapillary database to find pairs of sequences covering a wide range of geographical locations and appearances. The pairs are selected to satisfy the following criteria: 
\begin{enumerate*}
  \item Sequence length between 40 and 60 images.
  \item Minimum sequence density of 0.2 images/m.
  \item Maximum sequence-to-sequence distance of 3m.
  \item Each sequence can be reconstructed individually using SfM.
\end{enumerate*}

Note that we deliberately chose to focus on small scenes rather than attempting to collect data for a larger spatial area.
In practice, pose priors such as GPS are used to limit the search space during localization. 
Rather than artificially making the problem harder by ignoring such priors, we are interested in finding realistic hard examples with small scenes. 
Our dataset might not be as large as previous benchmarks. 
However, a large fraction of test images in previous benchmarks can be localized accurately. 
In contrast, our dataset contains many very challenging scenarios that current state-of-the-art methods cannot handle.

After collecting the sequence pairs, we run SfM on each of them. We assign a difficulty rating to each pair depending on the result of SfM and the relative orientation of the sequences:
Sequence pairs whose joint reconstruction succeeds\footnote{All images are included in the reconstruction with a sufficiently large number of well-distributed inliers}
are categorized as \textbf{easy} and are shown in ~Fig.~\ref{fig:geodist} (top).

If the joint reconstruction fails, we look at the average view direction of the sequences to categorize them into \textbf{medium}~(Fig.~\ref{fig:geodist} - left) if they have similar orientations (less than a 45$^\circ$ difference), or \textbf{hard}~(Fig.~\ref{fig:geodist} - right) if they have different orientations (more than a 45$^\circ$ difference).

This preliminary categorization is based on the following  observations: 
sequences with similar orientations are difficult to match mostly due to changes in appearance (\eg illumination or seasonal change) and state-of-the-art methods such as~\cite{sarlin2019coarse,sarlin2020superglue,Sarlin2021CVPR} are able to handle such changes quite well. 
In contrast, sequence pairs with different view directions are harder to match due to the addition of dramatic appearance changes of scene parts due 
the large viewpoint change and low visual overlap~\cite{schonberger2018semantic}. 
As our experiments in Sec.~\ref{sec:experiments} show, this preliminary classification aligns well with how current state-of-the-art methods perform on our dataset.

Tab.~\ref{tab:summary} shows statistics over the sequence pairs in CrowdDriven.
Many capture conditions that are challenging for current localization algorithms are included in CrowdDriven.
The medium category consists of different scene types shown under conditions such as day-night changes, rain, snow and other seasonal variations that impact the scene geometry.
The main focus of the hard category is the combination of 
large viewpoint changes with 
less extreme variations in illumination, seasons, and weather conditions.

\noindent \textbf{On the dataset size.} Similar to 
existing datasets~\cite{Sattler2018CVPR,Taira2018CVPR,wald-eccv-2020}, CrowdDriven only provides reference images taken under a single condition for each scene with test images.\footnote{The extended CMU and RobotCar Seasons provide training images taken under multiple conditions for a set of locations that does not overlap with the scene parts depicted in the test images.} 
As shown in Tab.~\ref{tab:dataset_comparison}, CrowdDriven contains fewer images than most such datasets, \eg, the CMU and RobotCar Seasons datasets from~\cite{Sattler2018CVPR}. 
However, this does not imply that our dataset is too small for learning-based methods such as camera pose~\cite{Kendall2015ICCV,Kendall2017CVPR,Walch2017ICCV,Brahmbhatt2018CVPR} or scene coordinate~\cite{Shotton2013CVPR,Brachmann2016CVPR,Brachmann2017CVPR,Brachmann2020ARXIV,Cavallari2017CVPR,Brachmann2019ICCV,Cavallari20193DV} regression. 
These approaches regress the camera pose from a given image and a 3D point coordinate from an image patch, respectively.
Both regression tasks are instance-level problems. 
As such, the number of images in local scene parts is more important for their performance in these parts than the absolute number of images in a dataset. 
Extending CrowdDriven in terms of the absolute number of images by adding additional scenes is easy. 
However, doing so is unlikely to improve the performance in existing scenes. 
Extending CrowdDriven by adding more images to existing scenes is hard as such images are simply not available: 
while crowd-sourced image capture allows us to obtain images from a diverse set of scenes from multiple continents, there is no control over how much data is acquired per scene. 
Still, the number of reference images per scene should be sufficient for 
learning-based 
techniques.

\begin{table*}[t]
\centering
\resizebox{12cm}{!}{
\begin{tabular}{|c|c|c|c|c|c|c|c|c|c|}
\toprule
scene type   & identifier & CP reproj. err. &  pos. STD (m) &  \# test\ images & \# ref.\ images. &  ref.\ conditions & test conditions & considerable changes & foliage  \\
\midrule
\rowcolor{LightGray} road  & Sydney &                -  &           -  &                   14  &                          28  &                       day, partly cloudy &      day, rain, &               illumination  &            \\                                                               
\rowcolor{LightGray}       & Massachusetts1 &                -  &           -  &          25  &                          49  &              day, partly cloudy &       day, overcast, &  illumination         & \checkmark \\                                                                         
\rowcolor{LightGray}       & Poing &                -  &           -  &                     20  &                          56  &                       day, clear sky &                 day, overcast  & illumination            & \checkmark \\                                                      
\rowcolor{LightGray}       & Washington &                -  &           -  &                10  &                          29  &                  day, clear sky &                   day, cloudy  & illumination          &            \\                                                             
\rowcolor{LightGray}       & Melbourne &                -  &           -  &                 12  &                          36  &                   day, cloudy &                   day, overcast &  illumination     &                \\                                                              
\rowcolor{MediumGray}       & Burgundy2 &    0.03\% (0.21 px)  &          0.07  &           50  &                          50  &             day, sunny &            day, rain &  illumination, rain &                       \\                                                                       
 \rowcolor{MediumGray}      & Thuringia &    0.07\% (0.46 px)  &          0.13  &           11  &                          17  &             day, sunny &                 day, clear sky & illumination   &          \checkmark  \\                                                                   
 \rowcolor{MediumGray}      & Massachusetts2 &    0.07\% (0.45 px)  &          0.07  &      24  &                          35  &        day, overcast &                     night                 & day-night   &                \\                                                                   
\rowcolor{DarkGray}       & Besançon2 &    0.01\% (0.07 px)  &          0.06  &             50  &                          50  &               day, overcast &             day, cloudy       & illumination, strong viewpoint & \checkmark     \\                                                     
      \rowcolor{DarkGray} & Besançon4 &    0.02\% (0.13 px)  &          0.06  &             50  &                          50  &               day, cloudy &          day, overcast & strong viewpoint, illumination & \checkmark   \\                                                                
      \rowcolor{DarkGray} & Besançon3 &    0.04\% (0.24 px)  &          0.04  &             50  &                          50  &               day, overcast &            day, vegetated,  strong viewpoint & illumination  & \checkmark \\                                                           
      \rowcolor{DarkGray} & Brittany &    0.11\% (0.69 px)  &          0.19  &            31  &                          53  &                day, sunny &            day, partly cloudy & strong viewpoint &  \checkmark                            \\                                               
      \midrule                                                                                                                                                                                                                                                                                        
\rowcolor{LightGray} suburban & Portland &                -  &           -  &            21  &                          41  &                 day, clear sky &                day, overcast &  illumination  &                       \\                                                               
      \rowcolor{LightGray} & Curitiba &                -  &           -  &               19  &                          20  &                    day, cloudy &             day, overcast & illumination  &                              \\                                                            
      \rowcolor{LightGray} & Tsuru &                -  &           -  &                  9  &                          26  &                       day, cloudy &          day, overcast & illumination    &                               \\                                                          
      \rowcolor{LightGray} & Clermont-Ferrand &                -  &           -  &        15  &                          21  &            day, sunny &                     day, overcast & illumination &                        \\                                                                   
      \rowcolor{LightGray} & Savannah &                -  &           -  &                18  &                          56  &                    day, clear sky &                day, cloudy   &  illumination  &                  \\                                                                
      \rowcolor{LightGray} & Subcarpathia &                -  &           -  &             17  &                          32  &                day, cloudy &                    day, overcast   & snow, seasonal &                                  \\                                                
      \rowcolor{MediumGray} & Massachusetts3 &    0.03\% (0.17 px)  &          0.04  &     44  &                          56  &              day, clear sky &                night              & day-night &            \checkmark      \\                                                           
      \rowcolor{MediumGray} & Skåne &    0.02\% (0.10 px)  &          0.03  &             20  &                          24  &                 day, cloudy &            day           & small viewpoint, illumination      &        \\                                                                
      \rowcolor{DarkGray} & Angers2 &    0.04\% (0.26 px)  &          0.05  &             46  &                          47  &                 day, cloudy &            day, clear sky  & strong viewpoint, illumination   &      \\                                                                  
      \rowcolor{DarkGray} & Ile-de-France &    0.02\% (0.13 px)  &          0.04  &        50  &                          50  &           day, sunny &                day,  cloudy         & strong viewpoint, illumination  &    \\                                                                  
      \rowcolor{DarkGray} & Orleans2 &    0.04\% (0.22 px)  &          0.05  &             31  &                          31  &                day, clear sky &                   day, sunny  & strong viewpoint, illumination     &       \\                                                         
      \rowcolor{DarkGray} & Pays de la Loire &    0.05\% (0.32 px)  &          0.03  &     42  &                          58  &              day, cloudy &                day,   overcast     & strong viewpoint & \checkmark          \\                                                             
      \rowcolor{DarkGray} & Brourges &    0.05\% (0.34 px)  &          0.06  &             22  &                          23  &                day, partly cloudy &     day,  clear sky, illumination   & strong viewpoint & \checkmark     \\                                                        
      \rowcolor{DarkGray} & Nouvelle-Aquitaine2 &    0.05\% (0.34 px)  &          0.05 &    45  &                          46  &              day, sunny &               day, sunny &  strong viewpoint    &        \\                                                                                
      \midrule                                                                                                                                                                                                                                                                                        
\rowcolor{LightGray}urban  & Muehlhausen &                -  &           -  &              10  &                          22  &                 day, cloudy &                 day, overcast &   slight illumination &   \\                                                                            
      \rowcolor{LightGray} & Bayern &                -  &           -  &                   26  &                          26  &                      day, cloudy &             day, overcast & illumination &        \\                                                                               
      \rowcolor{MediumGray} & Boston5 &    0.04\% (0.27 px)  &          0.08  &             34  &                          47  &               day, sunny &                 night      &  day-night &            \\                                                                                   
      \rowcolor{MediumGray} & Boston1 &    0.04\% (0.27 px)  &          0.05  &             47  &                          53  &               day, sunny &                 night      &  day-night &  \\                                                                                             
      \rowcolor{MediumGray} & Boston3 &    0.12\% (0.78 px)  &          0.52  &             31  &                          40  &               day, clear sky &           day,clear sky    &  small viewpoint  &        \\                                                                            
      \rowcolor{MediumGray} & Massachusetts4 &    0.06\% (0.40 px)  &          0.06  &      39  &                          45  &             day, clear sky &            night           &      day-night         & \\                                                                                
      \rowcolor{MediumGray} & Boston2 &    0.03\% (0.19 px)  &          0.03  &             49  &                          51  &               day, clear sky &           night            &      day-night         &  \\                                                                             
      \rowcolor{MediumGray} & Boston4 &    0.04\% (0.27 px)  &          0.05  &             34  &                          41  &               night &                    day, sunny       & day-night              &  \\                                                                             
      \rowcolor{MediumGray} & Cambridge &    0.11\% (0.69 px)  &          0.05  &           33  &                          35  &             night  &                     day, sunny       & day-night              & \\                                                                              
      \rowcolor{DarkGray} & Le-Mans &    0.04\% (0.28 px)  &            0.04  &             49  &                          51  &                 day, overcast &           day, overcast & illumination, strong viewpoint & \\                                                                        
      \rowcolor{DarkGray} & Nouvelle-Aquitaine1 &    0.06\% (0.40 px)  &          0.05   & 50  &                          50  &             day, overcast &            day,  overcast  & strong viewpoint,  seasonal, snow &  \\                                                                      
      \rowcolor{DarkGray} & Angers1 &    0.08\% (0.51 px)  &          0.08  &             47  &                          49  &                 day, cloudy &              day,  clear sky &  strong viewpoint &     \\                                                                                
      \rowcolor{DarkGray} & Orleans1 &    0.05\% (0.29 px)  &          0.11  &             33  &                          34  &                day, sunny &               day,  clear sky  &  strong viewpoint, illumination &    \\                                                                  
       \rowcolor{DarkGray} & Leuven &    0.08\% (0.50 px)  &          0.04  &             20  &                          21  &                  day, cloudy &              day, overcast & strong viewpoint &       \\                                                                                                                                                    
\bottomrule
\end{tabular}
}
\caption{Statistics of CrowdDriven: scene type, identifier, median reprojection error of control points (CPs), number of test and reference images, reference and test conditions and changes. Preliminary categorization: easy: light gray; medium: gray; hard: dark gray.}
\label{tab:summary}
\end{table*}

\vspace{-3pt}
\subsection{Reference Pose Generation}
\label{sec:dataset:reference_pose_generation}
\vspace{-3pt}
To use the sequences for benchmarking visual localization algorithms, two processing steps are required: 
\begin{enumerate*}
    \item estimating intrinsic camera calibrations and reference poses 
    for all images in a common coordinate system, such that distances can be measured in meters.
    \item subdividing the datasets into reference (training) images and test images. 
\end{enumerate*}
For (2), we use a simple strategy: 
for each pair of sequences, the larger sequence defines the reference images, and the images from the smaller sequence are used for testing.

\noindent \textbf{Easy Datasets} 
are those where both sequences can be reconstructed in a common coordinate frame by off-the-shelf SfM using SIFT~\cite{Lowe04IJCV} features. 

We follow common practice~\cite{Sattler2018CVPR,Frahm10ECCV,Kendall2015ICCV,Li2012ECCV} and use the camera poses and intrinsics estimated during the reconstruction process to define our reference poses. 
After running SfM on the sequences, we visually inspect the 3D models. 
If the two sequences are \emph{aligned}, \ie, there are no duplicate 3D points, and the camera poses look visually correct, we scale the SfM poses based on available GPS data to (approximately) recover the scale of the scene. 
For this category, we expect state-of-the-art localization approaches to perform well given that SIFT features are sufficiently powerful to register the sequences.

We used OpenSfM\cite{OpenSfM} as our SfM pipeline. 
Experiments with 
COLMAP\cite{Schoenberger2016CVPR} on our data showed similar results. 
Tab.~\ref{tab:summary} provides statistics over the 13 sequence pairs in the easy category.  
As can be seen, diversity in scenes such as weather and illumination changes is covered in different environments such as roads, suburban and urban areas.
Fig.~\ref{fig:geodist} (top) shows sample images from this category.

\begin{figure*}[t]
    \centering
    \subfloat{\includegraphics[width=0.28\textwidth, height=4cm]{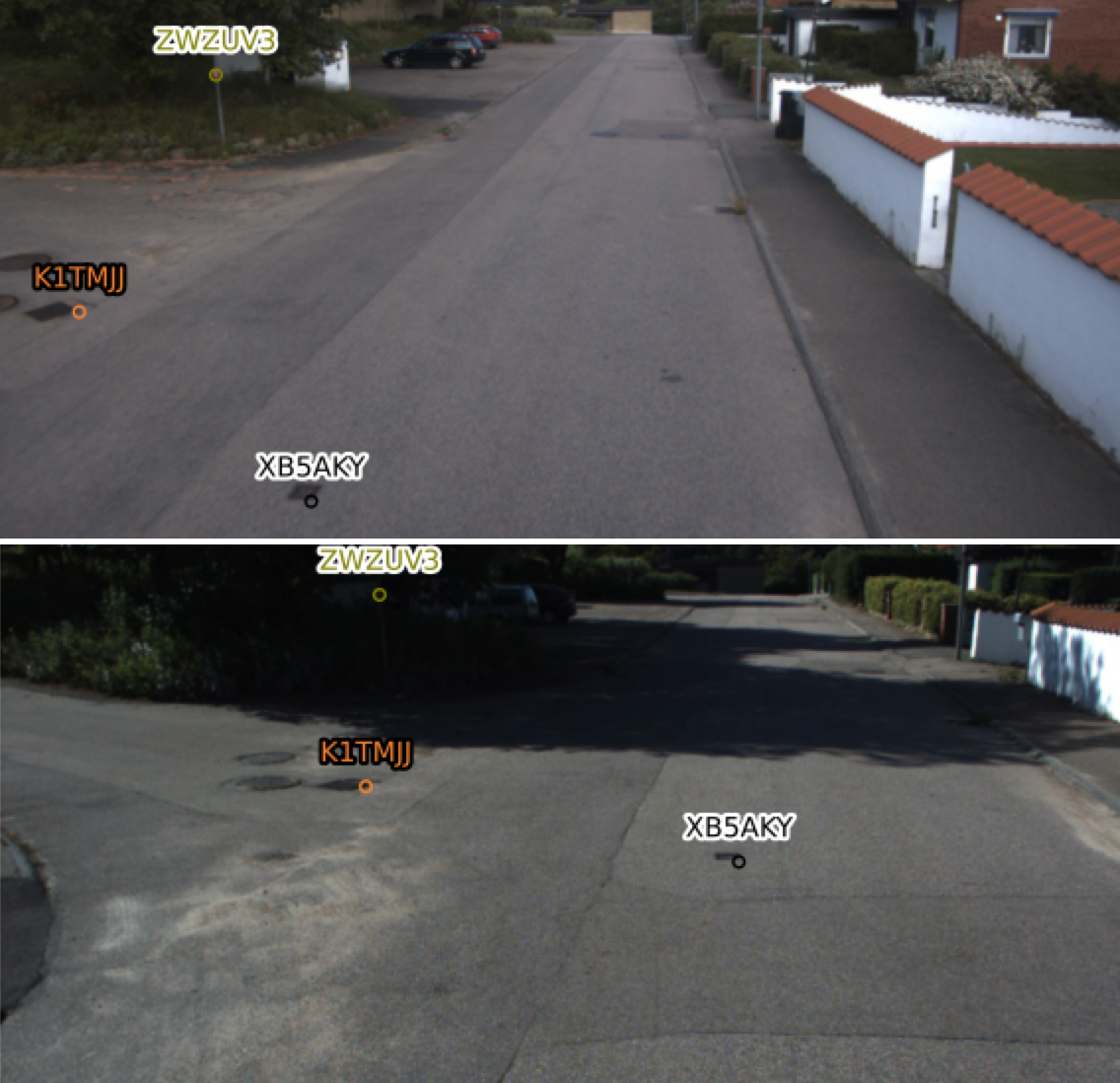}}
    
    \subfloat{\includegraphics[width=0.32\textwidth, height=4cm]{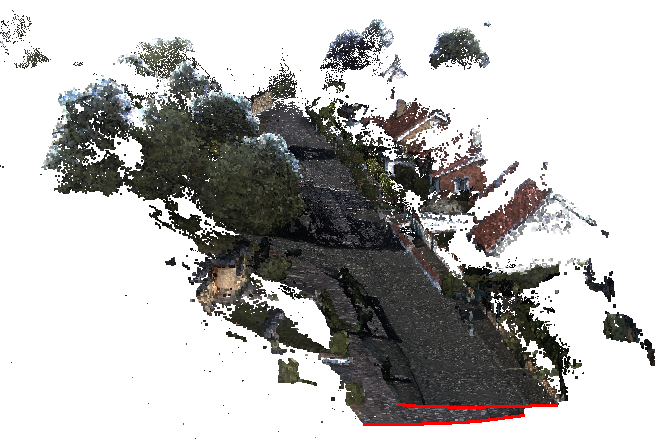}}
    
    \subfloat{\includegraphics[width=0.32\textwidth, height=4cm]{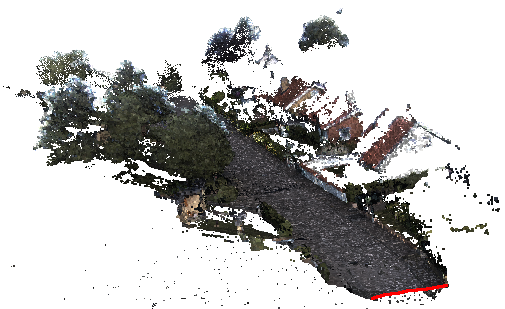}}
    
    \caption{Left: Annotated correspondences; Middle: Initial dense reconstruction of the scene; Right: Dense reconstruction with refinement after registration. The edge of the road surface is highlighted to indicate misalignment.}
    \label{fig:pose_refinement}
\end{figure*}

\begin{figure}[ht]
    \centering
    \includegraphics[width=0.7\columnwidth]{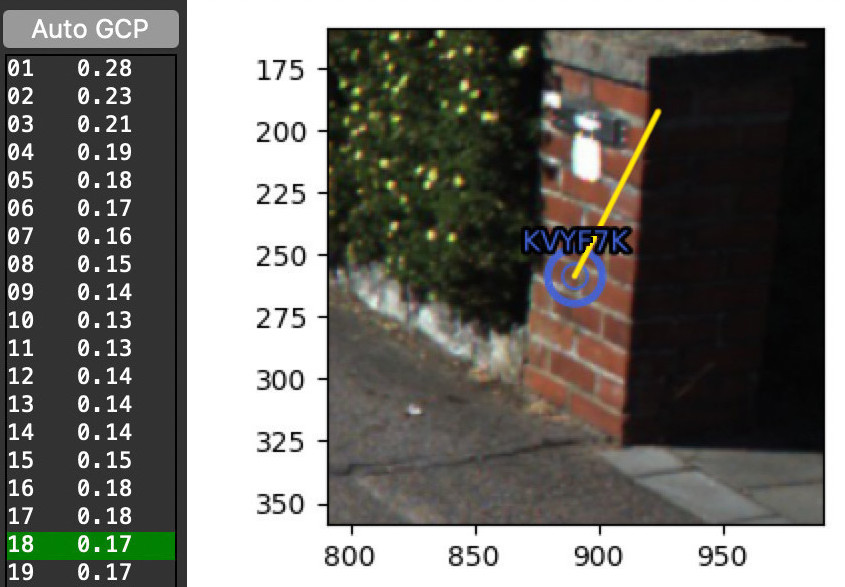}
    \caption{Our annotation UI integrates annotation metrics to enable efficient annotation and QA: (1) Reprojection error of annotated CPs (yellow line) (2) Positional std. deviation derived from bundle adjustment, highlighted in green
    (17 cm for this example).
    }
    \label{fig:ui}
\end{figure}

\noindent \textbf{Medium and Hard Datasets} 
are those where SfM fails to register the sequence pairs 
due to large changes in appearance (medium) and/or viewpoint (hard). 
Both medium and hard datasets are thus 
significantly more challenging 
for existing visual localization algorithms. 
Thus, these datasets will be most interesting to the community.

To obtain reference poses, we first reconstruct each sequence individually using SfM and recover the scale of the models using known GPS data from Mapillary. 
Next, we manually annotate corresponding pixel positions between images from the two sequences (Figs.~\ref{fig:pose_refinement}~\&~\ref{fig:ui}).
These annotations define manual tracks within each sequence and between sequences, which we will call control points (CPs). 
On average, 12 different control points have been annotated for each dataset (sequence pair), where each such point has been observed on average in 10 images.

To align the sequences, we first obtain the 3D position of the CPs in the reference frame of each reconstruction by triangulating the 2D annotations.
The corresponding 3D points (as per the annotation) provide us with 3D-3D matches that we use to compute an initial registration as a similarity transformation, bringing both reconstructions into a common reference frame.
Given this initial alignment, we perform bundle adjustment\cite{triggs1999bundle} over all camera poses and intrinsics, the 3D points triangulated from SIFT features, and the 3D points triangulated from annotations. 

Fig.~\ref{fig:pose_refinement} shows the quality of the alignment when using only GPS constraints (middle) and when using manual annotations (right). 
For clarity of visualization, we show the scenes as dense Multi-View Stereo point clouds, although the registration is performed using only sparse points.

We do not use dense point clouds for alignment as they cannot always be obtained solely from images, \eg, due to over- (bright sun) or underexposure (night) or lack of texture. Moreover, they are expensive to compute and introduce some drawbacks, \eg, registration failures due to missing or changed geometry (foliage, snow, \etc).
\vspace{-3pt}
\subsection{Quality Control}
\vspace{-3pt}
To verify the reference poses, we use the following criteria and workflow. If SfM succeeds and there are no noticeable artefacts and errors, then it typically produces highly accurate estimates for the camera poses and the 3D structure of the scene. 
Thus, we directly trust the reference poses we obtained for the easy datasets. 

For the medium and hard datasets, we rely on manually-annotated matches, 
which are susceptible to human error and might be distributed in a sub-optimal way in the scene, under-constraining the alignment. 

Previous work on building benchmark datasets that also uses manual annotations~\cite{Sattler2018CVPR,Taira2018CVPR} often relied on visual inspection and did not rigorously measure the uncertainty of their generated poses. 
In the case of the Aachen Day-Night dataset~\cite{Sattler2018CVPR,Sattler2012BMVC}, \cite{Zhang2020IJCV} recently showed that inaccurate poses passed visual inspection. 
We thus made a conscious effort to quantitatively measure the accuracy of our poses. 

We use two metrics to verify that the annotations are correct and sufficient. 
After a few control points have been annotated, an alignment is generated by estimating a similarity transform followed by bundle adjustment (as described above).
To detect potential issues, we begin by estimating the \textbf{reprojection error} of every annotation.
This metric identifies wrongly annotated points such as wrong correspondences between different objects with similar appearance or simply misclicks during annotation. All annotations with an error of more than one pixel (at VGA resolution) are flagged as incorrect and must be refined.

After all the reprojection errors are sufficiently small, 
we compute the \textbf{camera position covariance} for each image using bundle adjustment~\cite{ForstnerWrobel}.
To fix the gauge ambiguity, we fix the poses of one of the sequences, run the bundle adjustment problem, and compute the covariance of the poses of the other sequence.  We repeat this while fixing the poses of the other sequence.  From the covariances, we compute the standard deviation of the camera positions\footnote{
    We focus on the positional covariance and not the full pose covariance as (1) it is easier to understand and (2) the positions are less certain than the orientation estimates (which can be constrained by points at infinity)~\cite{ForstnerWrobel}.
} in meters and seek for small values when annotating.

This metric is used to understand if the calculated position of an image is under-constrained: 
all annotations can be correct with very small reprojection errors, but the poses might still be under-constrained, \eg, if the annotations were only performed on far-away points, resulting in large standard deviations. 

\noindent \textbf{Annotation Tools.} These two metrics are integrated with the annotation UI that we have developed and will release as part of this work (\cf Fig.~\ref{fig:ui}). Annotators can run bundle adjustment and get feedback directly in the UI, visualize wrongly annotated points and focus their efforts on annotating those frames with higher positional uncertainty, simplifying and accelerating the annotation process. Using our tool, the manual annotation (and QA) of a sequence pair reconstruction takes, on average, only 30 minutes.

After all positional std. deviations are confirmed to be under 30~cm, we perform a final check by visually inspecting the combined point cloud for the aligned models (Fig.~\ref{fig:pose_refinement}). Tab.~\ref{tab:summary} shows some statistics about the annotation process, including the number of annotated points, the median reprojection error and median positional std. deviation for each dataset. For the whole set of Medium and Hard datasets, the median annotation reprojection error is 0.28~px and the median positional std. deviation is 5.4~cm.

\vspace{-3pt}
\section{Baselines}
\vspace{-3pt}
In order to show that our benchmark introduces new challenges for localization algorithms, we use a set of state-of-the-art localization methods as baselines. 
We focus on methods that have been shown to work well under changing conditions (based on their performance on the \href{https://www.visuallocalization.net/benchmark/}{benchmark} from~\cite{Sattler2018CVPR}) and provide source code and trained models:

\noindent \textbf{HLoc~\cite{sarlin2019coarse,sarlin2020superglue}} 
uses learned SuperPoint~\cite{DeTone2018CVPRW} features and
SuperGlue~\cite{sarlin2020superglue} to establish 2D-3D matches with a SfM model, which are then used for camera pose estimation.
\noindent \textbf{D2-Net~\cite{dusmanu2019d2}} uses a single CNN for feature detection and description. 

Pose estimation is implemented in COLMAP~\cite{Schoenberger2016CVPR}. 

In contrast to HLoc, which runs (close to) real-time, D2-Net  requires multiple seconds per test image.

\noindent \textbf{Rectified SIFT~\cite{toft2020single}} uses 
a depth estimation network~\cite{MPSD_2020_ECCV} to detect planar regions in the image. Warping them to remove perspective foreshortening leads to features that can be matched under strong viewpoint changes.

\noindent \textbf{S2DHM~\cite{germain2019sparse}} uses an asymmetric matching approach: reference images are represented by sparse features corresponding to the 3D points in the SfM model, while the test image is represented by densely extracted descriptors.

\noindent The above-mentioned baselines are based on matching local features to establish 2D-3D matches between the test image and an SfM model of the scene. 
In contrast, \textbf{PixLoc}~\cite{Sarlin2021CVPR} does not rely on feature matching but refines an initial pose estimate by minimizing a feature-metric cost function.

We do not evaluate camera pose and scene coordinate regression methods. 
Pose regressors have been shown to be significantly less accurate than other localization approaches~\cite{Sattler2019CVPR}, 
even on scenes without changing conditions. 
We thus see no reason why they should perform well on our more challenging benchmark. 

Current scene coordinate regressors seem to struggle with strong condition changes between the training and test sets: 

on the Aachen Day-Night dataset~\cite{Sattler2012BMVC,Sattler2018CVPR}, where all training images are taken at day, ESAC~\cite{Brachmann2019ICCV} performs significantly worse for nighttime test images compared to daytime queries and is less accurate than HLoc by a wide margin.

CrowdDriven only provides references images taken under a single condition for each scene.
Given the strong condition changes between training and test images,
current scene coordinate regressors are unlikely to 
perform well on our dataset. 

In our experiments, for a given test image, we consider only reference images from the same scene, \eg, Boston1, and not from different scenes. 
Since our scenes are rather small (\cf Tab.~\ref{tab:summary}), using place recognition / image retrieval is not necessary. 
Instead, we exhaustively match the test image against all reference images in the scene.
\begin{table*}[t]
\resizebox{\linewidth}{!}{
\begin{tabular}{|c|c|c|c|c|c|c|c|c|c|c|c|c|c|c|c|c|c|c|c|}
\toprule
 & \multicolumn{3}{c}{D2-Net} & \multicolumn{3}{|c}{S2DHM} & \multicolumn{3}{|c}{HLoc} &  \multicolumn{3}{|c}{Rectified SIFT} & \multicolumn{3}{|c|}{PixLoc} & \\ \cmidrule{2-16}
                name &   pos. err &  rot. err &                   \thead{\% of localized \\ 0.5/1.0/5.0/10.0  (m) \\ 2/5/10/20 (\si{\degree})} &   pos. err &   rot. err &                         \thead{\% of localized \\ 0.5/1.0/5.0/10.0  (m) \\ 2/5/10/20 (\si{\degree})} &   pos. err &   rot. err &                         \thead{\% of localized \\ 0.5/1.0/5.0/10.0  (m) \\ 2/5/10/20 (\si{\degree})} &   pos. err &   rot. err &                         \thead{\% of localized \\ 0.5/1.0/5.0/10.0  (m) \\ 2/5/10/20 (\si{\degree})} & pos. err &   rot. err &                         \thead{\% of localized \\ 0.5/1.0/5.0/10.0  (m) \\ 2/5/10/20 (\si{\degree})} & Changes      \\
\midrule

\rowcolor{DarkGray} Angers1  &  28.02   &   177.43   &          F   &  97.81   & 171.26   &          F   &  46.39   & 161.65   &          F   & 191.62   & 148.27   &          F    &   21.03   &   175.11   &   F  &\includegraphics[width=14px]{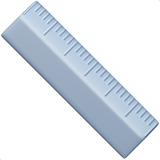} \\
\rowcolor{DarkGray} Angers2  &  35.51   &   165.56   &          F   & 174.61   & 153.78   &          F   &  68.34   & 122.82   &          0/ 0/ 0/ 6.52   & 437.63   & 132.38   &          F   &   45.26   &   173.18   &   F   &\includegraphics[width=14px]{emojis/stv.png} \\
\rowcolor{LightGray} Bayern  &   0.03   &     0.06   &      96.15/ 96.15/ 96.15/ 96.15   &   0.09   &   0.30   &      80.77/ 80.77/ 80.77/ 80.77   &   0.02   &   0.07   &      80.77/ 80.77/ 80.77/ 80.77   &   0.04   &   0.11   &      80.77/ 84.62/ 84.62/ 84.62  &   0.09   &   0.12   &   61.54/ 61.54/ 65.38/ 73.08 &\includegraphics[width=14px]{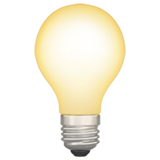} \includegraphics[width=14px]{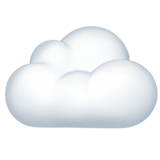}\includegraphics[width=14px]{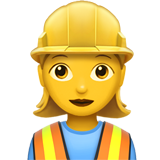}  \\
\rowcolor{DarkGray} Besançon2  &  81.45   &   160.22   &          F   &    -   &    -   &                          -   &  59.02   & 152.43   &          F   & 128.70   & 121.79   &          F    &   34.30   &   169.03   &   F  &\includegraphics[width=14px]{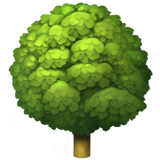} \includegraphics[width=14px]{emojis/stv.png} \\
\rowcolor{DarkGray} Besançon3  &  48.16   &   162.30   &          F   & 258.71   & 148.52   &          F   &  71.61   & 162.18   &          F   & 107.51   & 148.63   &          F   &   36.86   &   168.47   &   F  &\includegraphics[width=14px]{emojis/fo.png} \includegraphics[width=14px]{emojis/stv.png} \\
\rowcolor{DarkGray} Besançon4  & 117.25   &   151.57   &          F   &    -   &    -   &                          -   & 108.59   & 141.24   &          F   & 287.57   & 134.02   &          F   &   69.80   &   172.84   &   F  &\includegraphics[width=14px]{emojis/ov.png} \includegraphics[width=14px]{emojis/fo.png} \includegraphics[width=14px]{emojis/stv.png} \\
\rowcolor{DarkGray} Boston1  &  28.78   &     4.99   &         0/ 0/ 4.26/ 23.40   & 239.57   & 140.05   &          F   &  31.75   &   8.47   &         0/ 0/ 0/ 10.64   & 125.13   & 129.79   &          F  &     27.09   &     15.55   &     0/ 0/ 0/ 2.13 &\includegraphics[width=14px]{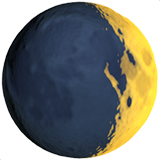}\includegraphics[width=14px]{emojis/mm.png}  \\
\rowcolor{MediumGray} Boston2  &   6.46   &     0.96   &        0/ 0/ 24.49/ 97.96   & 496.16   &  86.71   &         0/ 0/ 8.16/ 16.33   &   4.68   &   0.82   &        0/ 0/ 63.27/ 95.92   &  87.21   & 150.42   &          F    &   13.26   &   7.22   &   0/ 0/ 6.12/ 38.78 &\includegraphics[width=14px]{emojis/ng.png}\includegraphics[width=14px]{emojis/mm.png}  \\
\rowcolor{MediumGray} Boston3  &   6.66   &     4.20   &        0/ 0/ 29.03/ 51.61   & 196.98   & 114.24   &          F   &  27.57   &  32.83   &        0/ 0/ 12.90/ 19.35   & 111.70   & 155.53   &          F  &   20.72   &   16.12   &   0/ 0/ 0/ 19.35 &\includegraphics[width=14px]{emojis/il.png}\includegraphics[width=14px]{emojis/mm.png}  \includegraphics[width=14px]{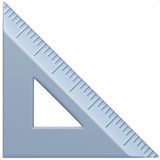} \\
\rowcolor{MediumGray} Boston4  &  12.94   &     2.51   &        0/ 0/ 20.83/ 41.67   &    -   &    -   &                          -   &  15.57   &   6.19   &        0/ 0/ 26.47/ 38.24   &    -   &    -   &                          -  &   18.90   &   5.81   &   0/ 0/ 11.76/ 26.47  &\includegraphics[width=14px]{emojis/ng.png}\includegraphics[width=14px]{emojis/mm.png}  \\
\rowcolor{MediumGray} Boston5  &  18.08   &     2.26   &        0/ 0/ 11.76/ 17.65   &  97.70   &  74.40   &          0/ 0/ 0/ 5.88   &  16.59   &   4.35   &        0/ 0/ 26.47/ 26.47   &  91.74   & 157.72   &          F   &   13.36   &   12.36   &   0/ 0/ 0/ 26.47 &\includegraphics[width=14px]{emojis/ng.png}\includegraphics[width=14px]{emojis/mm.png}  \\
\rowcolor{DarkGray} Brittany  &  14.74   &   147.24   &          F   & 177.38   & 143.03   &          F   &  36.44   & 137.46   &          0/ 3.23/ 3.23/ 3.23   & 305.68   & 121.02   &          F   &   14.96   &   162.08   &   F  &\includegraphics[width=14px]{emojis/fo.png}
\includegraphics[width=14px]{emojis/mm.png} \includegraphics[width=14px]{emojis/stv.png} \\
\rowcolor{DarkGray} Brourges  &  31.84   &   153.74   &          F   &    -   &    -   &                          -   &  22.24   & 153.54   &          F   &  57.10   &  97.23   &          F   &   14.64   &   177.38   &   F  &\includegraphics[width=14px]{emojis/fo.png} \includegraphics[width=14px]{emojis/stv.png}
\includegraphics[width=14px]{emojis/mm.png}  \\
\rowcolor{MediumGray} Burgundy2  &   4.41   &     3.72   &        0/ 4.00/ 60/ 76.00   &  57.78   &  34.59   &          F   &   7.32   &   5.36   &        0/ 4.00/ 40/ 58.00   & 554.19   & 166.83   &          F &   20.97   &   14.82   &   0/ 0/ 0/ 26.00 &\includegraphics[width=14px]{emojis/il.png} \includegraphics[width=14px]{emojis/ov.png} \includegraphics[width=14px]{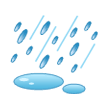}\\
\rowcolor{LightGray} Cambridge  &   0.50   &     0.87   &      51.52/ 90.91/ 93.94/ 96.97   &  94.58   &  82.57   &       9.09/ 12.12/ 12.12/ 12.12   &   0.37   &   0.43   &   69.70/ 100/ 100/ 100   &  58.58   & 135.56   &        3.03/ 6.06/ 18.18/ 21.21   &   30.06   &   15.24   &   0/ 0/ 3.03/ 6.06  &\includegraphics[width=14px]{emojis/ng.png}
\includegraphics[width=14px]{emojis/mm.png} \\
\rowcolor{LightGray} Clermont-Ferrand  &   0.22   &     0.24   &  100/ 100/ 100/ 100   &   0.25   &   0.47   &  100/ 100/ 100/ 100   &   0.15   &   0.27   &  100/ 100/ 100/ 100   &   0.21   &   0.33   &      80/ 93.33/ 93.33/ 93.33    &   0.19   &   0.20   &   93.33/ 93.33/ 93.33/ 93.33  &\includegraphics[width=14px]{emojis/il.png} \includegraphics[width=14px]{emojis/ov.png}
\includegraphics[width=14px]{emojis/mm.png} \\
\rowcolor{LightGray} Curitiba  &   0.03   &     0.07   &  100/ 100/ 100/ 100   &   0.21   &   0.36   &    84.21/ 89.47/ 100/ 100   &   0.04   &   0.06   &  100/ 100/ 100/ 100   &   0.06   &   0.08   &      89.47/ 89.47/ 89.47/ 89.47      &   0.06   &   0.09   &   84.21/ 84.21/ 84.21/ 84.21  &\includegraphics[width=14px]{emojis/il.png} \includegraphics[width=14px]{emojis/ov.png}
\includegraphics[width=14px]{emojis/mm.png} \\
\rowcolor{DarkGray} Ile-de-France  &  58.17   &   159.89   &          F   &    -   &    -   &                          -   & 121.89   & 120.76   &          F   & 325.41   & 159.15   &          F    &   24.99   &   175.11   &   F  &\includegraphics[width=14px]{emojis/stv.png}
\includegraphics[width=14px]{emojis/mm.png} \\
\rowcolor{DarkGray} Le-Mans  &  53.53   &   160.64   &          F   &  63.90   & 165.43   &          F   &  52.82   & 163.53   &          F   & 249.61   & 138.46   &          F   &   40.68   &   176.24   &   F  &\includegraphics[width=14px]{emojis/ov.png} \includegraphics[width=14px]{emojis/stv.png}
\includegraphics[width=14px]{emojis/mm.png} \\
\rowcolor{DarkGray} Leuven  &  10.85   &   164.76   &          F   &  48.83   & 154.03   &          F   &  12.54   & 173.48   &          F   &  69.29   & 132.33   &          F    &   8.42   &   140.80   &   F  &\includegraphics[width=14px]{emojis/ov.png} \includegraphics[width=14px]{emojis/stv.png}
\includegraphics[width=14px]{emojis/mm.png} \\
\rowcolor{LightGray} Massachusetts1  &   0.08   &     0.06   &  100/ 100/ 100/ 100   &   0.59   &   0.36   &      40/ 72.00/ 96.00/ 96.00   &   0.12   &   0.07   &  100/ 100/ 100/ 100   &   0.16   &   0.06   &    96.00/ 96.00/ 100/ 100     &   0.20   &   0.10   &   76.00/ 84.00/ 84.00/ 84.00 &\includegraphics[width=14px]{emojis/il.png} \includegraphics[width=14px]{emojis/ov.png} \includegraphics[width=14px]{emojis/fo.png}
\includegraphics[width=14px]{emojis/mm.png} \\
\rowcolor{DarkGray} Massachusetts2  &   5.80   &     0.24   &        0/ 0/ 0/ 100   &    -   &    -   &                          -   &  18.34   &   9.47   &          0/ 0/ 0/ 8.33   & 111.63   & 168.17   &          F   &   6.70   &   10.76   &   0/ 0/ 0/ 66.67  &\includegraphics[width=14px]{emojis/ng.png}
\includegraphics[width=14px]{emojis/mm.png} \\
\rowcolor{MediumGray} Massachusetts3  &  23.95   &    25.32   &       0/ 10.53/ 36.84/ 42.11   & 498.52   & 123.61   &          F   &   3.91   &   5.85   &       0/ 25.00/ 52.27/ 59.09   & 680.80   & 104.46   &          F     &   18.02   &   10.24   &   0/ 0/ 4.55/ 22.73 & \includegraphics[width=14px]{emojis/fo.png} \includegraphics[width=14px]{emojis/ng.png}
\includegraphics[width=14px]{emojis/mm.png} \\
\rowcolor{MediumGray} Massachusetts4  &   1.57   &     1.01   &       0/ 0/ 97.44/ 100   &  69.55   &  24.08   &        0/ 8.11/ 32.43/ 32.43   &   2.23   &   1.14   &      0/ 0/ 100/ 100   &  40.25   & 115.17   &          0/ 0/ 0/ 4.00    &   10.38   &   5.70   &   0/ 0/ 30.77/ 46.15 &\includegraphics[width=14px]{emojis/ng.png}
\includegraphics[width=14px]{emojis/mm.png} \\
\rowcolor{LightGray} Melbourne  &   0.07   &     0.07   &  100/ 100/ 100/ 100   &   0.21   &   0.22   &   83.33/ 100/ 100/ 100   &   0.09   &   0.16   &  100/ 100/ 100/ 100   &   0.16   &   0.16   &  100/ 100/ 100/ 100     &   14.07   &   0.93   &   16.67/ 16.67/ 16.67/ 41.67 &\includegraphics[width=14px]{emojis/il.png}
\includegraphics[width=14px]{emojis/mm.png} \\
\rowcolor{LightGray} Muehlhausen  &   0.03   &     0.07   &  100/ 100/ 100/ 100   &   0.32   &   0.53   &   80/ 100/ 100/ 100   &   0.04   &   0.07   &  100/ 100/ 100/ 100   &   0.04   &   0.11   &  100/ 100/ 100/ 100   &   0.04   &   0.10   &   100/ 100/ 100/ 100 &\includegraphics[width=14px]{emojis/ov.png}
\includegraphics[width=14px]{emojis/mm.png} \\
\rowcolor{DarkGray} Nouvelle-Aquitaine1  &  40.23   &   172.90   &          F   & 328.30   & 150.36   &          F   &  34.02   & 160.82   &          F   & 418.74   & 137.36   &          F   &   44.17   &   170.42   &   F   &\includegraphics[width=14px]{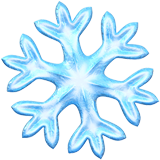} \includegraphics[width=14px]{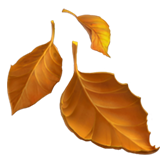} \includegraphics[width=14px]{emojis/stv.png}\\
\rowcolor{DarkGray} Nouvelle-Aquitaine2  &  81.48   &   126.82   &          F   &  35.89   & 161.41   &          F   &  67.65   & 147.96   &          F   &  90.10   & 150.32   &          F   &   28.51   &   169.94   &   F  &\includegraphics[width=14px]{emojis/stv.png} \\
\rowcolor{DarkGray} Orleans1  &  17.42   &   178.86   &          F   & 256.48   & 149.29   &          F   &  33.42   & 175.46   &          F   &  87.40   & 157.44   &          0/ 0/ 3.03/ 3.03   &   25.87   &   178.74   &   F  &\includegraphics[width=14px]{emojis/stv.png} \\
\rowcolor{DarkGray} Orleans2  & 175.99   &   127.58   &          F   &    -   &    -   &                          -   &  32.95   & 165.35   &          F   & 359.37   & 154.26   &          F  &   15.30   &   177.55   &   F  &\includegraphics[width=14px]{emojis/stv.png}
\includegraphics[width=14px]{emojis/mm.png} \\
\rowcolor{DarkGray} Pays de la Loire  &  25.32   &   159.23   &          F   &  52.74   & 156.08   &          F   &  34.11   & 166.72   &          F   & 131.31   & 135.25   &          0/ 0/ 0/ 4.76   &   17.56   &   175.95   &   F  &\includegraphics[width=14px]{emojis/fo.png} \includegraphics[width=14px]{emojis/stv.png}
\includegraphics[width=14px]{emojis/mm.png} \\
\rowcolor{LightGray} Poing  &   0.05   &     0.07   &  100/ 100/ 100/ 100   &   0.45   &   0.61   &    60/ 85.00/ 100/ 100   &   0.06   &   0.07   &  100/ 100/ 100/ 100   & 183.46   &  85.13   &      20/ 20/ 20/ 20.00    &   0.08   &   0.04   &   85.00/ 85.00/ 85.00/ 85.00 &\includegraphics[width=14px]{emojis/il.png} \includegraphics[width=14px]{emojis/fo.png}
\includegraphics[width=14px]{emojis/mm.png} \\
\rowcolor{LightGray} Portland  &   0.13   &     0.16   &  100/ 100/ 100/ 100   &   0.40   &   0.46   &    66.67/ 95.24/ 100/ 100   &   0.11   &   0.14   &  100/ 100/ 100/ 100   &   0.16   &   0.12   &   95.24/ 100/ 100/ 100     &   0.13   &   0.16   &   85.71/ 85.71/ 85.71/ 85.71 &\includegraphics[width=14px]{emojis/il.png} \includegraphics[width=14px]{emojis/ov.png}
\includegraphics[width=14px]{emojis/mm.png} \\
\rowcolor{LightGray} Savannah  &   0.08   &     0.05   &  100/ 100/ 100/ 100   &   0.25   &   0.28   &    83.33/ 94.44/ 100/ 100   &   0.08   &   0.05   &  100/ 100/ 100/ 100   &   0.07   &   0.05   &  100/ 100/ 100/ 100   &   0.09   &   0.08   &   94.44/ 94.44/ 94.44/ 94.44 &\includegraphics[width=14px]{emojis/il.png}
\includegraphics[width=14px]{emojis/mm.png} \\
\rowcolor{MediumGray} Skåne  &   5.14   &     2.97   &        0/ 0/ 50/ 85.00   & 392.36   & 120.41   &          0/ 0/ 0/ 5.00   &   4.26   &   3.70   &        0/ 5.00/ 55.00/ 90.00   & 609.69   & 154.13   &          F   &   35.93   &   25.73   &   F  &\includegraphics[width=14px]{emojis/il.png} \includegraphics[width=14px]{emojis/smv.png} \\
\rowcolor{LightGray} Subcarpathia  &   0.54   &     0.46   &      47.06/ 70.59/ 88.24/ 94.12   &   6.39   &   4.09   &        0/ 0/ 41.18/ 58.82   &   0.35   &   0.26   &    70.59/ 76.47/ 100/ 100   & 116.17   &  80.70   &      11.76/ 11.76/ 29.41/ 35.29   &   66.44   &   9.84   &   5.88/ 5.88/ 11.76/ 11.76  &\includegraphics[width=14px]{emojis/sn.png} \includegraphics[width=14px]{emojis/se.png}
\includegraphics[width=14px]{emojis/mm.png} \\
\rowcolor{LightGray} Sydney  &   0.18   &     0.11   &  100/ 100/ 100/ 100   &   1.39   &   0.82   &      35.71/ 35.71/ 85.71/ 92.86   &   0.18   &   0.15   &   85.71/ 100/ 100/ 100   &   2.81   &   1.86   &       0/ 25.00/ 75.00/ 75.00    &   0.18   &   0.20   &   64.29/ 71.43/ 71.43/ 71.43 &\includegraphics[width=14px]{emojis/il.png} \includegraphics[width=14px]{emojis/ov.png}
\includegraphics[width=14px]{emojis/mm.png} \\
\rowcolor{LightGray} Thuringia  &   0.57   &     0.26   &    45.45/ 90.91/ 100/ 100   &   0.95   &   0.60   &    18.18/ 54.55/ 100/ 100   &   0.37   &   0.25   &   81.82/ 100/ 100/ 100   & 431.41   & 121.14   &          F     &   7.03   &   6.51   &   0/ 0/ 27.27/ 63.64  &\includegraphics[width=14px]{emojis/il.png}
\includegraphics[width=14px]{emojis/mm.png} \\
\rowcolor{LightGray} Tsuru  &   0.01   &     0.04   &  100/ 100/ 100/ 100   &   0.06   &   0.30   &  100/ 100/ 100/ 100   &   0.03   &   0.03   &  100/ 100/ 100/ 100   &   0.02   &   0.04   &  100/ 100/ 100/ 100   &    0.03   &   0.04   &   100/ 100/ 100/ 100 & \includegraphics[width=14px]{emojis/il.png} \includegraphics[width=14px]{emojis/ov.png}
\includegraphics[width=14px]{emojis/mm.png} \\
\rowcolor{MediumGray} Washington  &   1.00   &     0.42   &     0/ 50/ 100/ 100   &   3.96   &   0.66   &      0/ 0/ 100/ 100   &   1.07   &   1.47   &    10/ 30/ 100/ 100   &   0.90   &   0.68   &     40/ 50/ 90/ 100   &   2.46   &   1.20   &   0/ 0/ 70.00/ 70.00  &\includegraphics[width=14px]{emojis/il.png} \\
\bottomrule
\end{tabular}
}

\caption{Localization performance 
on our CrowdDriven benchmark. We report the median position (in meters) and orientation (in degrees) errors, and the percentage of test images localized within certain error bounds on the position and orientation errors. Easy, medium, and hard datasets are color-coded in light, standard, and dark gray, respectively. We also provide information about the type of change between the training and test sequences: illumination: 
\protect\includegraphics[width=8px]{emojis/il.png}, 
overcast: \protect\includegraphics[width=8px]{emojis/ov.png}, foliage: \protect\includegraphics[width=8px]{emojis/fo.png}, snow: \protect\includegraphics[width=8px]{emojis/sn.png}, seasonal: \protect\includegraphics[width=8px]{emojis/se.png}, day-night: \protect\includegraphics[width=8px]{emojis/ng.png}, small viewpoint: \protect\includegraphics[width=8px]{emojis/smv.png} , rain: \protect\includegraphics[width=8px]{emojis/rn.png}, strong viewpoint: \protect\includegraphics[width=8px]{emojis/stv.png}, man-made changes:  \protect\includegraphics[width=8px]{emojis/mm.png} .
'F' stands for failure to localize any image within the coarsest precision regime.
}
\label{tab:all_baseline}
\end{table*}

The supp. material shows results for \textbf{sequence-based localization} methods 
that simultaneously estimate the poses of all images in a sequence~\cite{Kukelova2016CVPR,Sweeney2014ECCV,Ventura2014CVPR,Lee2015IJRR,Camposeco2016ECCV}. 

While this
improves performance, it is still not sufficient to obtain reasonable performance on our more challenging scenes.

\vspace{-3pt}
\section{Experimental Evaluation}
\label{sec:experiments}
\vspace{-3pt}

\begin{figure*}[]
    \centering
    \includegraphics[width =0.75\textwidth]{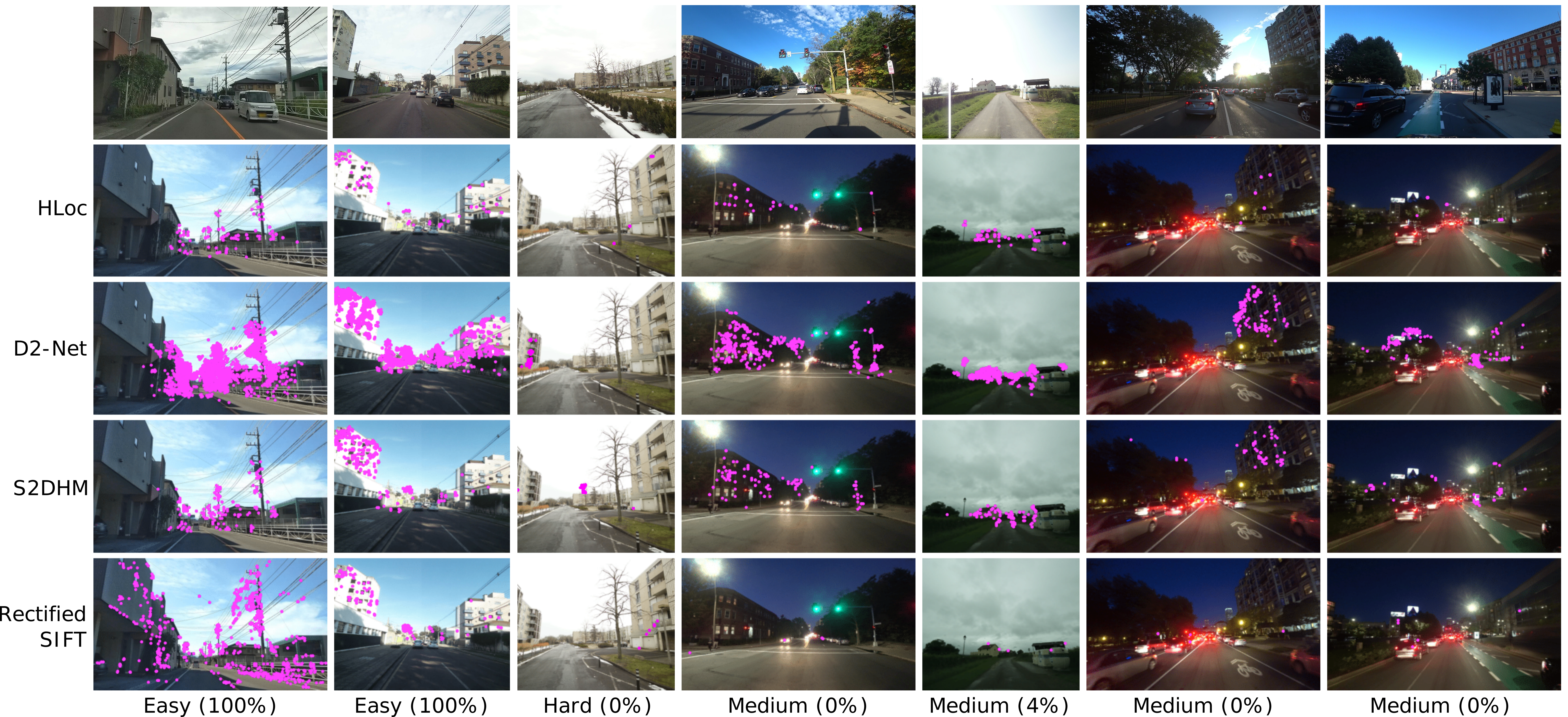}
    \caption{Inlier plots for selected scenes. 
    None of the evaluated methods 
    is able to robustly localize the \textit{medium} and \textit{hard} datasets. This is made apparent by the fraction of localized images under the \textit{medium-precision} threshold for the \textbf{best} method, shown in below each column.
    }
    \label{fig:inliers}
\end{figure*}

This section 
shows how our dataset introduces new challenging scenarios that most baselines cannot cope with. 
After introducing our evaluation measures, we analyze the impact of different types of changes on the performance.

\noindent \textbf{Evaluation Measures.}
We follow common evaluation protocols~\cite{Kendall2015ICCV,Frahm10ECCV,Sattler2018CVPR,Taira2018CVPR} and report median position (in meters) and orientation errors (in degrees), as well as the percentage of test images with poses that differ within certain error bounds from their reference poses. 
The position error is measured as the Euclidean distance between the reference and estimated position. 
To measure the orientation error, we compute the minimum rotation angle $\alpha$ that aligns the estimated rotation matrix $\mathtt{R}_\text{est}$ with the reference rotation $\mathtt{R}_\text{ref}$ as $2 cos(|\alpha|) = \text{trace}(\mathtt{R}^T_\text{ref} \mathtt{R}_\text{est}) - 1$~\cite{hartley2013rotation}. 

Inspired by~\cite{Sattler2018CVPR}, we use three error bounds: high~precision (images localized within 0.5m and 2\si{\degree} of their reference poses), medium~precision (1m, 5\si{\degree}), and coarse~precision (5m, 10\si{\degree}). 
In addition, we introduce a very~coarse~precision regime (10m, 20\si{\degree}).

We also report the changes in conditions between the test and training images: 
(slight) illumination changes (il.), \eg, between a cloudy and a sunny day, foliage / no foliage on the vegetation (fo.), snow / no snow on the ground (sn.), other seasonal changes (se.) (summer/autumn),
day-night changes (ng.), rain / no rain (rn.), small (sm. v.) and strong viewpoint (st. v.), and man-made (\eg, appearance / disappearance of cars) changes. Tab.~\ref{tab:all_baseline} summarizes the results obtained by evaluating the baselines.
In the following, we focus our analysis on three types of changes: \textbf{slight illumination}, \textbf{day-night}, and \textbf{strong viewpoint} changes, which are the dominant types of changes.

\noindent \textbf{Slight illumination changes.}
\noindent As shown in Tab.~\ref{tab:all_baseline}, datasets such as Muehlhausen, Tsuru, Poing, Bayern, Savannah, Curitiba, Melbourne, Sydney, and Clermont-Ferrand are categorized as being affected solely by illumination changes seen during the day.
Almost all the test images are localized with high precision by most of the methods. 
In this scenario, SIFT is already robust enough to slight changes in illumination. However, rectified SIFT  struggles to deal with the heavy vegetation present in Poing, Sydney, and Subcarpathia. 
We observe that for some datasets, PixLoc performs significantly worse than the other baselines. 
We attribute this to different viewing condition between the training set of PixLoc and our dataset, \eg, PixLoc was not trained on images showing as much snow as the Subcarpathia dataset, and the challenges of identifying the best pose from all available pose estimates (\cf supp. mat.) and of pose refinement in complex scenes such as Melbourne and Sydney. 
Still, we conclude that 
slight illumination changes are not challenging for current algorithms. 
\noindent \textbf{Day-night changes.}
The significant appearance changes that occur between day and night have considerable effects on the performance of most of the methods, leading to a noticeable increase in the errors. 
D2-Net’s robustness to illumination changes makes the model perform well in some of the datasets such as Massachusetts4 and Boston4, allowing localization with medium precision.
On the other hand, the effect of artificial lights during the night could cover the previously-seen features or considerably change them, making one part of the image sparser, and the pose estimation less accurate. In Massachusetts2, Massachusetts3, Boston2, and Boston5, the errors are larger than those of consumer-grade GPS. 
HLoc and S2DHM, as seen in Tab.~\ref{tab:all_baseline}, produce large errors in all of the night cases except for Cambridge, with the median errors of these methods starting at more than 30 meters.
We can conclude that for night-time test images, due to the considerable changes in feature appearances, 
 approaches such as S2DHM do not seem to improve the results as long as the extracted Hypercolumn descriptors themselves are not robust enough to the day-night changes (\cf Fig.~\ref{fig:inliers}). Rectified SIFT inherits the limitations of SIFT features and also struggles significantly on these scenes. 
\noindent \textbf{Strong viewpoint changes.} 
\noindent
By far the most challenging condition present in our dataset is strong viewpoint change,
as seen in
Tab.~\ref{tab:all_baseline}.
The rotation errors obtained for all the methods are over 160\si{\degree}, indicating complete failures. 
Strong viewpoint changes significantly alter the appearance of objects, resulting in substantially different feature descriptors depending on the viewing angle. 

In theory, removing perspective distortion via rectification should help to better handle larger viewpoint changes~\cite{toft2020single}. 
Yet, 
the rectified SIFT baseline still mostly fails under strong viewpoint changes. 
This can be attributed to the limited visual overlap as well as inaccurate depth predictions on the distant overlapping image regions (\cf Fig.~\ref{fig:inliers}). 
Unsurprisingly, PixLoc always fails under strong viewpoint changes:  
PixLoc refines an initial pose estimate, which is obtained from the poses of the database images. 
In the case of opposite viewpoints, these initial estimates are simply too different from the test poses for PixLoc to converge to a reasonable pose estimate. 

\noindent \textbf{Difficulty levels revisited.} 
Sec.~\ref{sec:dataset} provided a preliminary classification 
based on our understanding of what is challenging for current localization algorithms. 
Based on the measured performance of our baselines, we revisit this classification: 
a dataset should be considered as easy if (nearly) all baselines perform well, as medium if some baselines work well (at least for the coarsest threshold), and as hard if all methods fail. 
We color-coded the datasets in Tab.~\ref{tab:all_baseline} accordingly. 
Comparing the color-codings in Tabs.~\ref{tab:summary} and \ref{tab:all_baseline} shows that the classification based on performance is consistent with our preliminary categorization for most scenes.

\vspace{-4pt}
\section{Conclusion}
\vspace{-3pt}
We have introduced the CrowdDriven dataset for visual localization, with a focus on exposing the failure cases of state-of-the-art pipelines.
The dataset covers three difficulty tiers: (1) slight illumination changes, (2) severe illumination changes, (3) severe viewpoint changes.

We have evaluated the performance of the best visual localization methods available on our dataset and analyzed the failure modes for the different methods. We conclude that there is a large performance gap to be covered by future work and that can be evaluated with our dataset. 
Finally, we release the annotation tool we developed to create this dataset, enabling easy extension to more scenarios.

As shown concurrently to our work~\cite{Brachmann2021ICCV}, using a feature-based approach (SfM) to generate reference poses can create a bias towards feature-based methods. 
We consider the problem of eliminating such a bias from the evaluation as an important direction for future work.

{
\small{
\noindent \textbf{Acknowledgements.} This work has been funded by the Chalmers AI Research Centre (VisLocLearn), the Swedish Foundation for Strategic Research (Semantic Mapping and Visual Navigation for Smart Robots), the EU Horizon 2020 project RICAIP (grant agreement No 857306), and the European Regional Development Fund under IMPACT No.~CZ.02.1.01/0.0/0.0/15$\_$003/0000468.
}
}
\appendix
\section*{Appendix}

This appendix provides the following information: 
Sec.~\ref{sec:pixloc} provides details about the PixLoc baseline used in the main paper.
Sec.~\ref{sec:multi_image} provides a more detailed description of the sequence-based localization approaches used in this work and presents experimental results. 
Sec.~\ref{sec:dataset_supp} shows images from the scenes included in our proposed benchmark. 

\section{PixLoc Details}
\label{sec:pixloc}
Given an initial pose estimate, a set of database images, and a set of 3D points potentially visible in it, PixLoc~\cite{Sarlin2021CVPR} refines the initial estimate by minimizing a feature-metric error: 
PixLoc computes feature maps for the images, projects the 3D points into the test images and the database images, and minimizes the difference in the features belonging to the projections. 
For a test image, we consider each database image from the same scene. 
For each database image $\mathcal{I}_D$, we identify the four other database images taken from the most similar positions and use the 3D points visible in these five database images. 
The pose of $\mathcal{I}_D$ is used to initialize the pose of the test image. 
This results in $N$ poses for the test image, one for each of the $N$ database images. 
We tried selecting the pose with the smallest feature-metric error but found that this approach did not work well. 
Likely, this is due to comparing poses that observe different 3D points. 
Rather, we use a simple scoring function for each pose: 
let $\mathbf{c}_i$ be the test image position estimated for the $i$-th database image. 
The score $s_i$ for the $i$-th pose (computed from the $i$-th database image) is given as
\begin{equation}
    s_i = \sum_{j \not= j} \frac{1}{||\mathbf{c}_i - \mathbf{c}_j||_2 + \varepsilon} \enspace ,
\end{equation}
where $\varepsilon$ is a small constant to avoid division by zero. 
Intuitively, the score for a pose is large if there are multiple other pose estimates nearby. 
As such, the scoring is based on the assumption that we will have multiple pose estimates close to the true test pose while incorrect poses are rather far from each other. 

For PixLoc, we use the 3D models created for the HLoc method~\cite{sarlin2019coarse,sarlin2020superglue} as we got better results with these models compared to those build using SIFT~\cite{Lowe04IJCV} and D2-Net~\cite{dusmanu2019d2} features.

\section{Multi-image localization}
\label{sec:multi_image}
As discussed in the main paper, in addition to single-image localization, we have evaluated a multi-image (sequence-based) localization approach on our dataset. 
Using known relative poses, this approach models a sequence of images as a generalized camera~\cite{Pless2003CVPR}, \ie, a camera with multiple centers of projections. 
This enables us to estimate the poses of all images in the sequence at the same time~\cite{Kukelova2016CVPR,Sweeney2014ECCV,Ventura2014CVPR,Lee2015IJRR,Camposeco2016ECCV}. 
The advantage of this approach is that it enables localizing multiple images even if none of them individually has enough correct matches to facilitate successful pose estimation. 

We use the multi-image approach from~\cite{wald-eccv-2020}, using the code publicly released by the authors.\footnote{\url{https://github.com/tsattler/MultiCameraPose}}
This method use a minimal solver~\cite{Kukelova2016CVPR} (inside a LO-RANSAC~\cite{Lebeda2012BMVC} loop) that estimates both the pose of the generalized camera and its intrinsic scale, \ie, the scale of the distances between the individual images in a sequence. 
This models the fact that some approaches, \eg, monocular SLAM, might not be able to estimate the scale. 
Each new best model found inside RANSAC is optimized using local optimization~\cite{Lebeda2012BMVC}. 
This includes a non-linear optimization of the sum of squared reprojection errors over the inlier matches of the estimated generalized poses. 
Finally, the best pose found by RANSAC is optimized using the same optimization method. 
Non-linear optimization is implemented using the Ceres library~\cite{ceres-solver}. 
The 2D-3D matches required for estimating the pose of a generalized camera are provided by the baselines used in the main paper. 
\Ie, we use the 2D-3D matches found for each individual image in a generalized camera. 

We use sequences of length $k$ to define the generalized cameras. 
More precisely, for the $i$-th image in the sequence and a given sequence length $k$, we create a generalized camera containing images $i$ to $i+k$.\footnote{For images in the end of the sequence, the generalized camera might contain less than $k$ images.} 

As a result, each image is contained in multiple generalized cameras. 
Thus, there are multiple pose candidates for each test image, corresponding to the poses of the generalized cameras it is part of. 
We select the pose corresponding to the generalized camera pose with the largest number of inliers.

As in~\cite{Sattler2018CVPR}, we obtain the \emph{relative} poses required to form generalized cameras using the ground truth poses. 
This allows us to obtain an upper bound on the pose accuracy that can be obtained via multi-image queries. 
As can be seen from the results shown in Tab.~3 of the main paper, HLoc~\cite{sarlin2019coarse,sarlin2020superglue} and D2-Net~\cite{dusmanu2019d2} clearly outperform the Rectified SIFT~\cite{toft2020single} and S2DHM~\cite{germain2019sparse} baselines. 
In our experiments, we thus focus on HLoc and D2-Net. 
Note that we are not using PixLoc as the multi-image localization approach described above requires 2D-3D correspondences while PixLoc is not providing 2D-3D matches.

Table~\ref{tab:multi_image} provides detailed results on the easy (light gray), medium (gray), and hard (dark gray) parts of our benchmark. 
We use all images in a sequence to define the sequence length $k$. 
The results on the easy scenes are included as a form of sanity check to show that sequence-based localization works as intended. 
For reference, we also include Tab.~3 from the main paper, which provides single-image localization results, as Tab.~\ref{tab:all_baseline} here. 

As can be seen by comparing the two tables, using sequences typically improves the performance for the easy datasets, especially in terms of median position and orientation errors. 
For the medium datasets (standard gray), we observe that using sequences instead of single images for localization again improves results for many datasets. 
These improvements can be substantial, \eg, for Boston2, Boston4, Cambridge, Massachusetts3, and Thuringia. 
However, there are some exceptions, \eg, Boston1 and Burgundy2, where using sequences can actually lead to less accurate results. 
We attribute this to instabilities in the scale estimate refined by the local optimization, \eg, due to the distribution of 2D-3D matches over the images in the sequences. 
Looking at the hard datasets (dark gray), we observe that using sequence-based localization does not enable us to achieve significantly better results, even though we are using all images in a sequence and ground truth poses to define the generalized cameras. 
This result shows that our benchmark contains challenges that cannot be easily solved by simply using image sequences for localization.

\begin{table*}[tb]
\centering
\resizebox{13cm}{!}{
\begin{tabular}{|c|c|c|c|c|c|c|c|c|c|c|c|c|c|c|c|c|c|c|}
\toprule
 & \multicolumn{3}{c|}{D2-Net} & \multicolumn{3}{|c|}{HLoc} &   \\ \cmidrule{2-7}
                name &    pos. err &   rot. err &                         \thead{\% of localized \\ 0.5/1.0/5.0/10.0  (m) \\ 2/5/10/20 (\si{\degree})} &   pos. err &   rot. err &                         \thead{\% of localized \\ 0.5/1.0/5.0/10.0  (m) \\ 2/5/10/20 (\si{\degree})} &  Changes      \\
\midrule

\rowcolor{DarkGray} Angers1  &  36.62   &   179.54   &          F   &  31.98   & 178.82   &          F   &\includegraphics[width=14px]{emojis/stv.png} \\
\rowcolor{DarkGray} Angers2  &  70.23   &   162.00   &          F   &  11.20   &   5.10   &         0/ 0/6.52/ 45.65   &\includegraphics[width=14px]{emojis/stv.png} \\
\rowcolor{LightGray} Bayern  &   0.01   &     0.06   &  100/ 100/ 100/ 100   &   0.01   &   0.05   &  100/ 100/ 100/ 100   &\includegraphics[width=14px]{emojis/il.png} \includegraphics[width=14px]{emojis/ov.png}\includegraphics[width=14px]{emojis/mm.png} \\
\rowcolor{DarkGray} Besançon2  &  20.67   &   178.70   &          F   & 124.43   & 147.95   &          F   &\includegraphics[width=14px]{emojis/fo.png} \includegraphics[width=14px]{emojis/stv.png} \\
\rowcolor{DarkGray} Besançon3  &  43.82   &   173.62   &          F   &  69.34   & 162.69   &          F   &\includegraphics[width=14px]{emojis/fo.png} \includegraphics[width=14px]{emojis/stv.png} \\
\rowcolor{DarkGray} Besançon4  & 104.50   &   154.95   &          F   &  67.19   & 143.19   &          F   &\includegraphics[width=14px]{emojis/ov.png} \includegraphics[width=14px]{emojis/fo.png} \includegraphics[width=14px]{emojis/stv.png} \\
\rowcolor{DarkGray} Boston1  &  32.67   &     3.13   &          F   &  23.96   &   2.38   &          0/ 0/ 0/ 2.13   &\includegraphics[width=14px]{emojis/ng.png}\includegraphics[width=14px]{emojis/mm.png} \\
\rowcolor{MediumGray} Boston2  &   5.27   &     0.28   &       0/ 0/ 48.98/ 100   &   4.15   &   0.22   &       0/ 0/ 83.67/ 100   &\includegraphics[width=14px]{emojis/ng.png} \\
\rowcolor{MediumGray} Boston3  &   8.90   &     2.76   &        0/ 0/ 22.58/ 61.29   &   3.60   &   3.56   &        0/ 0/ 64.52/ 74.19   &\includegraphics[width=14px]{emojis/il.png} \includegraphics[width=14px]{emojis/smv.png}\includegraphics[width=14px]{emojis/mm.png} \\
\rowcolor{MediumGray} Boston4  &   4.66   &     1.40   &        2.94/ 2.94/ 55.88/ 91.18   &   4.61   &   1.99   &       5.88/ 14.71/ 64.71/ 88.24   &\includegraphics[width=14px]{emojis/ng.png}\includegraphics[width=14px]{emojis/mm.png} \\
\rowcolor{MediumGray} Boston5  &  10.73   &     2.28   &        0/ 0/ 26.47/ 47.06   &   2.34   &   1.80   &        0/ 0/ 67.65/ 67.65   &\includegraphics[width=14px]{emojis/ng.png}\includegraphics[width=14px]{emojis/mm.png} \\
\rowcolor{DarkGray} Brittany  &  59.55   &   142.10   &          F   &  20.51   & 144.92   &          0/ 0/ 9.68/ 9.68   &\includegraphics[width=14px]{emojis/fo.png} \includegraphics[width=14px]{emojis/stv.png}\includegraphics[width=14px]{emojis/mm.png} \\
\rowcolor{DarkGray} Brourges  &  22.36   &   175.96   &          F   &  24.28   & 172.75   &          F   &\includegraphics[width=14px]{emojis/fo.png} \includegraphics[width=14px]{emojis/stv.png}\includegraphics[width=14px]{emojis/mm.png} \\
\rowcolor{MediumGray} Burgundy2  &   4.15   &     4.18   &        0/ 0/ 58.00/ 96.00   &   8.05   &   3.39   &        0/ 0/ 30.00/ 62.00   &\includegraphics[width=14px]{emojis/il.png} \includegraphics[width=14px]{emojis/ov.png} \includegraphics[width=14px]{emojis/rn.png} \\
\rowcolor{LightGray} Cambridge  &   0.30   &     0.54   &   84.85/ 100/ 100/ 100   &   0.28   &   0.38   &  100/ 100/ 100/ 100   &\includegraphics[width=14px]{emojis/ng.png}\includegraphics[width=14px]{emojis/mm.png} \\
\rowcolor{LightGray} Clermont-Ferrand  &   0.17   &     0.20   &  100/ 100/ 100/ 100   &   0.15   &   0.10   &  100/ 100/ 100/ 100   &\includegraphics[width=14px]{emojis/il.png} \includegraphics[width=14px]{emojis/ov.png}\includegraphics[width=14px]{emojis/mm.png} \\
\rowcolor{LightGray} Curitiba  &   0.01   &     0.03   &  100/ 100/ 100/ 100   &   0.02   &   0.02   &  100/ 100/ 100/ 100   &\includegraphics[width=14px]{emojis/il.png} \includegraphics[width=14px]{emojis/ov.png}\includegraphics[width=14px]{emojis/mm.png} \\
\rowcolor{DarkGray} Ile-de-France  &  32.65   &   178.20   &          F   & 100.58   & 165.77   &          F   &\includegraphics[width=14px]{emojis/stv.png}\includegraphics[width=14px]{emojis/mm.png} \\
\rowcolor{DarkGray} Le-Mans  &  54.45   &   175.24   &          F   &  66.17   & 172.88   &          F   &\includegraphics[width=14px]{emojis/ov.png} \includegraphics[width=14px]{emojis/stv.png}\includegraphics[width=14px]{emojis/mm.png} \\
\rowcolor{DarkGray} Leuven  &  14.28   &   175.78   &          F   &  12.03   & 175.96   &          F   &\includegraphics[width=14px]{emojis/ov.png} \includegraphics[width=14px]{emojis/stv.png}\includegraphics[width=14px]{emojis/mm.png} \\
\rowcolor{LightGray} Massachusetts1  &   0.03   &     0.02   &  100/ 100/ 100/ 100   &   0.10   &   0.04   &  100/ 100/ 100/ 100   &\includegraphics[width=14px]{emojis/il.png} \includegraphics[width=14px]{emojis/ov.png} \includegraphics[width=14px]{emojis/fo.png}\includegraphics[width=14px]{emojis/mm.png} \\
\rowcolor{DarkGray} Massachusetts2  &   8.13   &     2.85   &        0/ 4.17/ 12.50/ 58.33   &  33.68   &  86.38   &         0/ 0/ 4.17/ 12.50   &\includegraphics[width=14px]{emojis/ng.png}\includegraphics[width=14px]{emojis/mm.png} \\
\rowcolor{MediumGray} Massachusetts3  &   2.34   &     2.72   &       0/ 18.18/ 86.36/ 95.45   &   2.03   &   3.41   &       0/ 34.09/ 95.45/ 97.73   &\includegraphics[width=14px]{emojis/fo.png} \includegraphics[width=14px]{emojis/ng.png}\includegraphics[width=14px]{emojis/mm.png} \\
\rowcolor{MediumGray} Massachusetts4  &   1.30   &     0.86   &      0/ 2.56/ 100/ 100   &   1.84   &   1.01   &       0/ 7.69/ 97.44/ 100   &\includegraphics[width=14px]{emojis/ng.png}\includegraphics[width=14px]{emojis/mm.png} \\
\rowcolor{LightGray} Melbourne  &   0.05   &     0.03   &  100/ 100/ 100/ 100   &   0.06   &   0.07   &  100/ 100/ 100/ 100   &\includegraphics[width=14px]{emojis/il.png}\includegraphics[width=14px]{emojis/mm.png} \\
\rowcolor{LightGray} Muehlhausen  &   0.02   &     0.07   &  100/ 100/ 100/ 100   &   0.03   &   0.09   &  100/ 100/ 100/ 100   &\includegraphics[width=14px]{emojis/ov.png}\includegraphics[width=14px]{emojis/mm.png} \\
\rowcolor{DarkGray} Nouvelle-Aquitaine1  &  52.11   &   173.29   &          F   &  55.75   & 176.71   &          F   &\includegraphics[width=14px]{emojis/sn.png} \includegraphics[width=14px]{emojis/se.png} \includegraphics[width=14px]{emojis/stv.png} \\
\rowcolor{DarkGray} Nouvelle-Aquitaine2  &  47.36   &   173.73   &          F   &  67.90   & 166.50   &          F   &\includegraphics[width=14px]{emojis/stv.png} \\
\rowcolor{DarkGray} Orleans1  &  30.21   &   179.61   &          F   &  38.52   & 179.27   &          F   &\includegraphics[width=14px]{emojis/stv.png} \\
\rowcolor{DarkGray} Orleans2  &  49.18   &   173.41   &          F   &  22.35   & 174.73   &          F   &\includegraphics[width=14px]{emojis/stv.png}\includegraphics[width=14px]{emojis/mm.png} \\
\rowcolor{DarkGray} Pays de la Loire  &  27.04   &   166.66   &          F   &  26.04   & 177.43   &          F   &\includegraphics[width=14px]{emojis/fo.png} \includegraphics[width=14px]{emojis/stv.png}\includegraphics[width=14px]{emojis/mm.png} \\
\rowcolor{LightGray} Poing  &   0.02   &     0.01   &  100/ 100/ 100/ 100   &   0.02   &   0.01   &  100/ 100/ 100/ 100   &\includegraphics[width=14px]{emojis/il.png} \includegraphics[width=14px]{emojis/fo.png}\includegraphics[width=14px]{emojis/mm.png} \\
\rowcolor{LightGray} Portland  &   0.04   &     0.04   &  100/ 100/ 100/ 100   &   0.06   &   0.03   &  100/ 100/ 100/ 100   &\includegraphics[width=14px]{emojis/il.png} \includegraphics[width=14px]{emojis/ov.png}\includegraphics[width=14px]{emojis/mm.png} \\
\rowcolor{LightGray} Savannah  &   0.03   &     0.01   &  100/100/100/100   &   0.02   &   0.02   &  100/100/100/100   &\includegraphics[width=14px]{emojis/il.png}\includegraphics[width=14px]{emojis/mm.png} \\
\rowcolor{MediumGray} Skåne  &  12.34   &     2.90   &        0/ 0/ 30.00/ 45.00   &   5.98   &   3.25   &        0/ 0/ 40.00/ 80.00   &\includegraphics[width=14px]{emojis/il.png} \includegraphics[width=14px]{emojis/smv.png}\includegraphics[width=14px]{emojis/mm.png} \\
\rowcolor{LightGray} Subcarpathia  &   0.30   &     0.09   &   82.35/ 100/ 100/ 100   &   0.32   &   0.08   &   82.35/ 100/ 100/ 100   &\includegraphics[width=14px]{emojis/sn.png} \includegraphics[width=14px]{emojis/se.png}\includegraphics[width=14px]{emojis/mm.png} \\
\rowcolor{LightGray} Sydney  &   0.04   &     0.02   &  100/ 100/ 100/ 100   &   0.04   &   0.03   &  100/ 100/ 100/ 100   &\includegraphics[width=14px]{emojis/il.png} \includegraphics[width=14px]{emojis/ov.png}\includegraphics[width=14px]{emojis/mm.png} \\
\rowcolor{LightGray} Thuringia  &   0.42   &     0.11   &    81.82/ 90.91/ 100/ 100   &   0.26   &   0.09   &  100/ 100/ 100/ 100   &\includegraphics[width=14px]{emojis/il.png}\includegraphics[width=14px]{emojis/mm.png} \\
\rowcolor{LightGray} Tsuru  &   0.01   &     0.04   &  100/ 100/ 100/ 100   &   0.02   &   0.07   &  100/ 100/ 100/ 100   &\includegraphics[width=14px]{emojis/il.png} \includegraphics[width=14px]{emojis/ov.png}\includegraphics[width=14px]{emojis/mm.png} \\
\rowcolor{MediumGray} Washington  &   0.69   &     0.36   &   10.00/ 100/ 100/ 100   &   1.00   &   0.87   &     0/ 50.00/ 100/ 100   &\includegraphics[width=14px]{emojis/il.png} \\
\bottomrule
\end{tabular}
}
\caption{Performance of the baseline methods on our CrowdDriven benchmark when using sequence-based localization. We report the median position (in meters) and orientation (in degrees) errors, as well as the percentage of test images localized within certain error bounds on the position and orientation errors. We report results for using all test images in a scene to define the generalized camera used for sequence-based localization. Easy, medium, and hard datasets are color-coded in light, standard, and dark gray, respectively. The right side of the table provides information about the type of change between the training and test sequences: illumination: 
\protect\includegraphics[width=11px]{emojis/il.png}, 
overcast: \protect\includegraphics[width=11px]{emojis/ov.png}, foliage: \protect\includegraphics[width=11px]{emojis/fo.png}, snow: \protect\includegraphics[width=11px]{emojis/sn.png}, seasonal: \protect\includegraphics[width=11px]{emojis/se.png}, day-night: \protect\includegraphics[width=11px]{emojis/ng.png}, small viewpoint: \protect\includegraphics[width=11px]{emojis/smv.png} , rain: \protect\includegraphics[width=11px]{emojis/rn.png}, strong viewpoint: \protect\includegraphics[width=11px]{emojis/stv.png}, man-made changes:  \protect\includegraphics[width=8px]{emojis/mm.png} .
'F' stands for failure to localize any image within the coarsest precision regime. }
\label{tab:multi_image}
\end{table*}

\section{Images from CrowdDriven}
\label{sec:dataset_supp}
\begin{figure}[tb] 
\centering  
\includegraphics[width=0.9\linewidth]{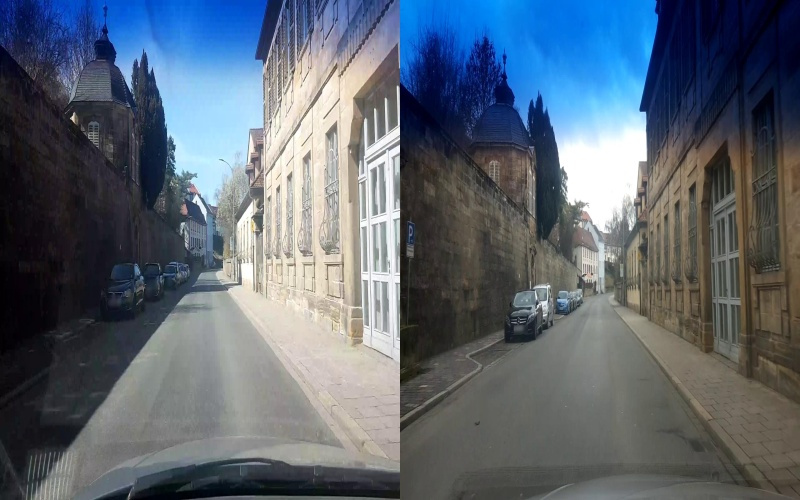}  
\caption{Bayern, Category : Easy}  
\label{fig:Bayern} 
\end{figure}

\begin{figure}[tb] 
\centering  
\includegraphics[width=0.9\linewidth]{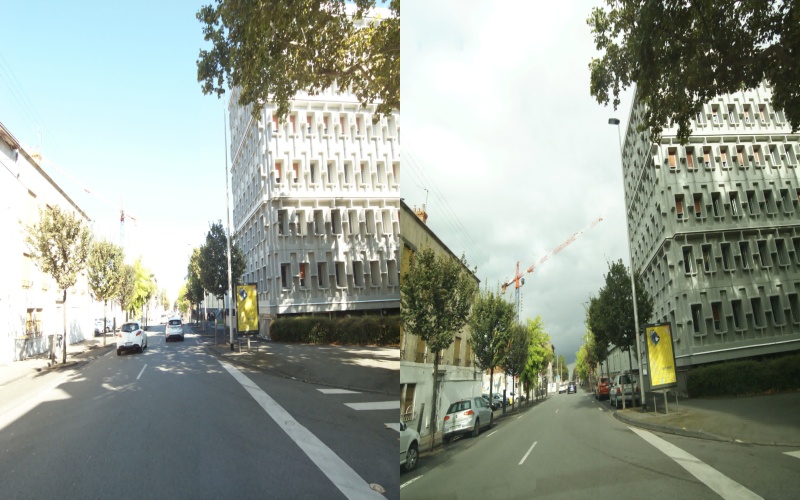}  
\caption{Clermont-Ferrand, Category : Easy}  
\label{fig:Clermont-Ferrand} 
\end{figure}

\begin{figure}[tb] 
\centering  
\includegraphics[width=0.9\linewidth]{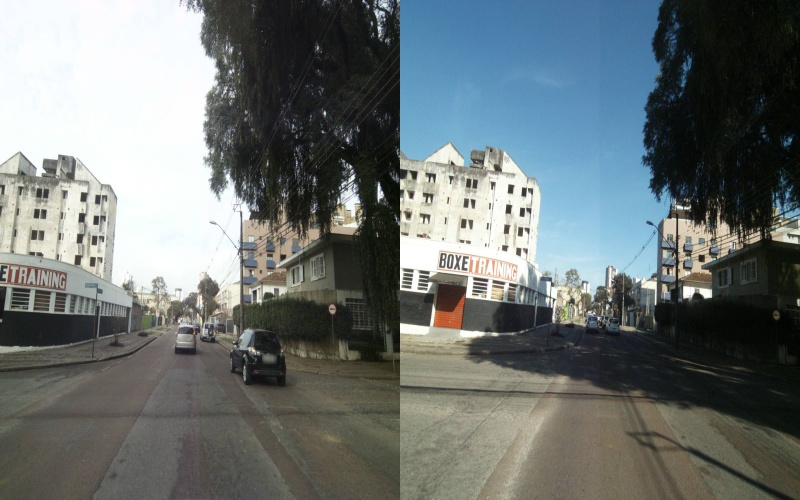}  
\caption{Curitiba, Category : Easy}  
\label{fig:Curitiba} 
\end{figure}

\begin{figure}[tb] 
\centering  
\includegraphics[width=0.9\linewidth]{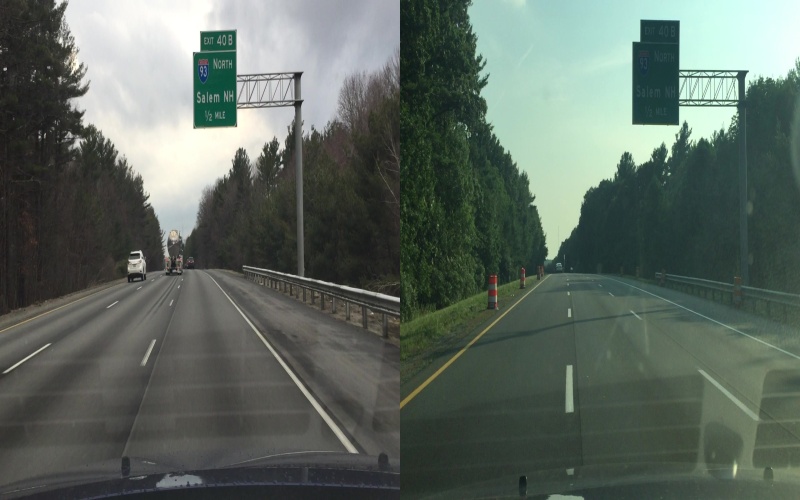}  
\caption{Massachusetts1, Category : Easy}  
\label{fig:Massachusetts1} 
\end{figure}

\begin{figure}[tb] 
\centering  
\includegraphics[width=0.9\linewidth]{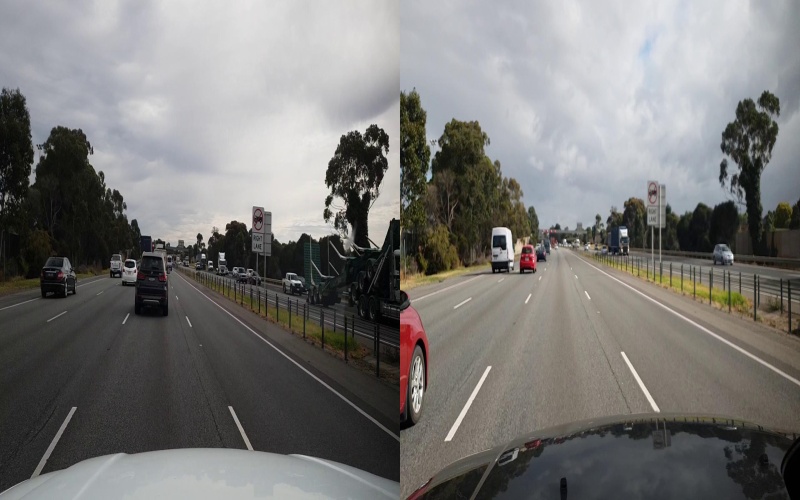}  
\caption{Melbourne, Category : Easy}  
\label{fig:Melbourne} 
\end{figure}

\begin{figure}[tb] 
\centering  
\includegraphics[width=0.9\linewidth]{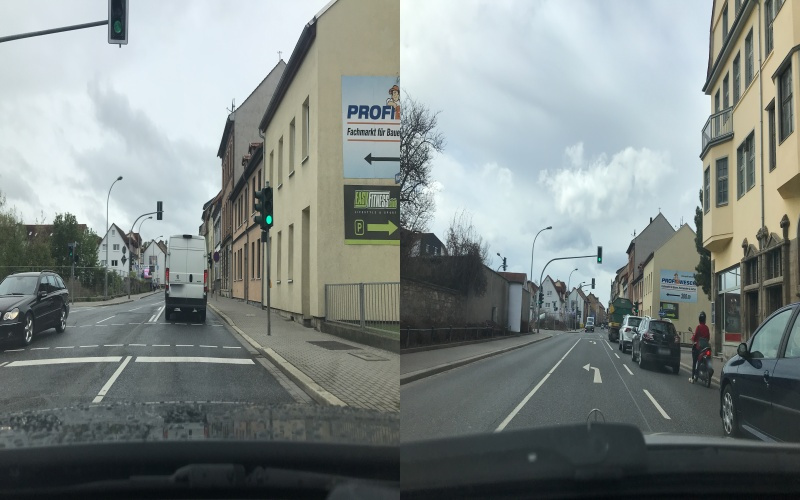}  
\caption{Muehlhausen, Category : Easy}  
\label{fig:Muehlhausen} 
\end{figure}

\begin{figure}[tb] 
\centering  
\includegraphics[width=0.9\linewidth]{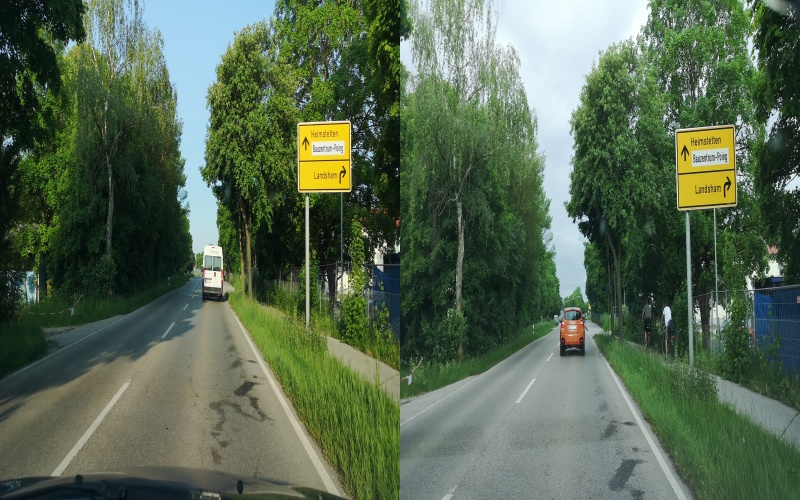}  
\caption{Poing, Category : Easy}  
\label{fig:Poing} 
\end{figure}

\begin{figure}[tb] 
\centering  
\includegraphics[width=0.9\linewidth]{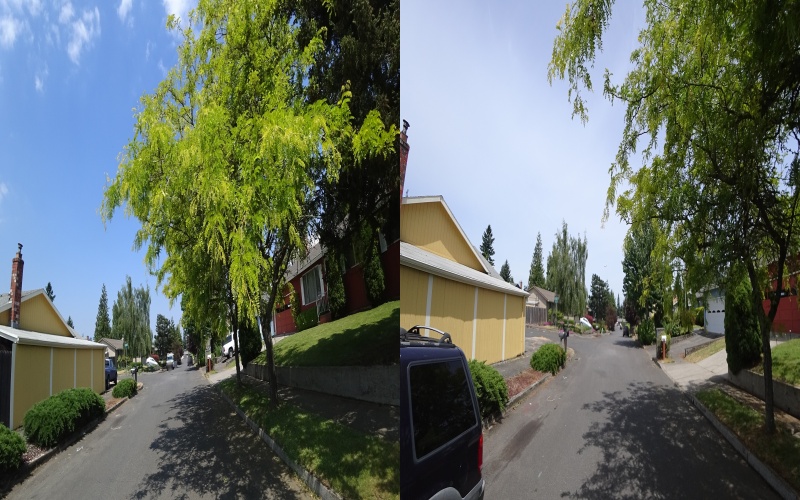}  
\caption{Portland, Category : Easy}  
\label{fig:Portland} 
\end{figure}

\begin{figure}[tb] 
\centering  
\includegraphics[width=0.9\linewidth]{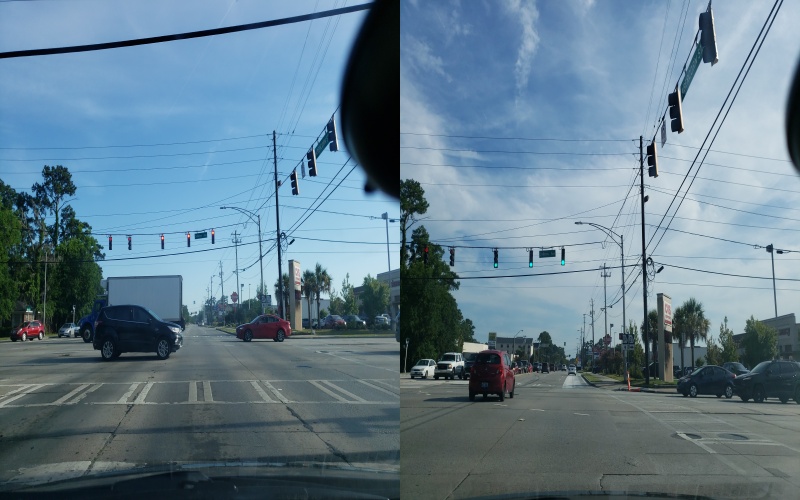}  
\caption{Savannah, Category : Easy}  
\label{fig:Savannah} 
\end{figure}

\begin{figure}[tb] 
\centering  
\includegraphics[width=0.9\linewidth]{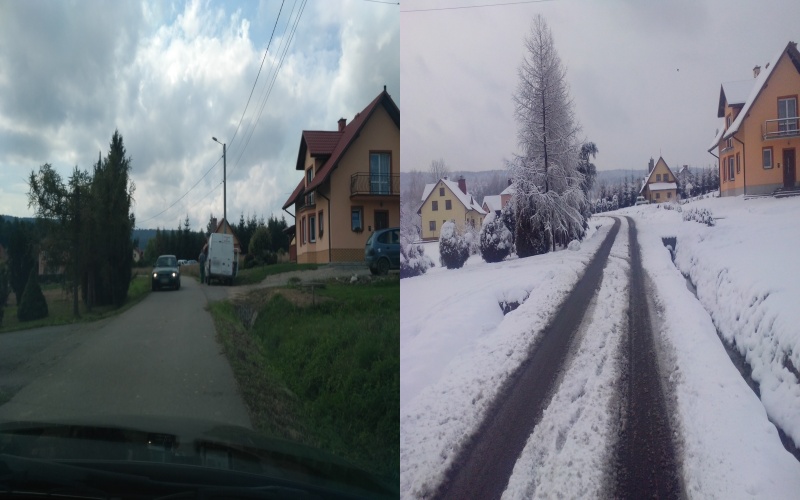}  
\caption{Subcarpathia, Category : Easy}  
\label{fig:Subcarpathia} 
\end{figure}

\begin{figure}[tb] 
\centering  
\includegraphics[width=0.9\linewidth]{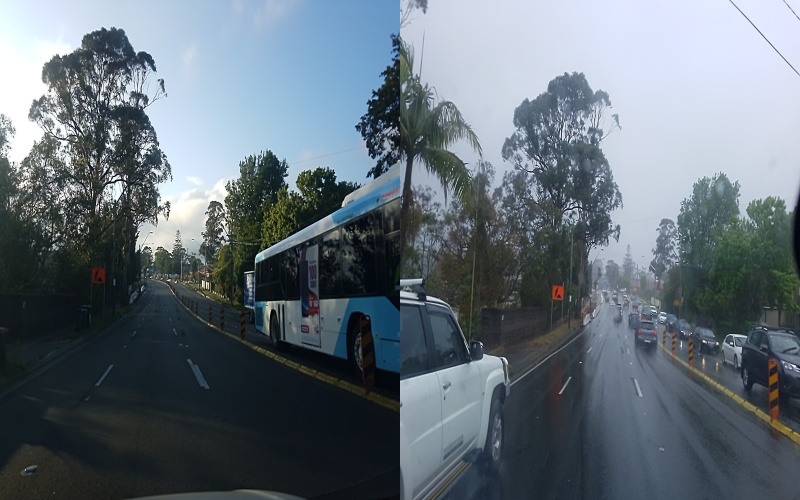}  
\caption{Sydney, Category : Easy}  
\label{fig:Sydney} 
\end{figure}

\begin{figure}[tb] 
\centering  
\includegraphics[width=0.9\linewidth]{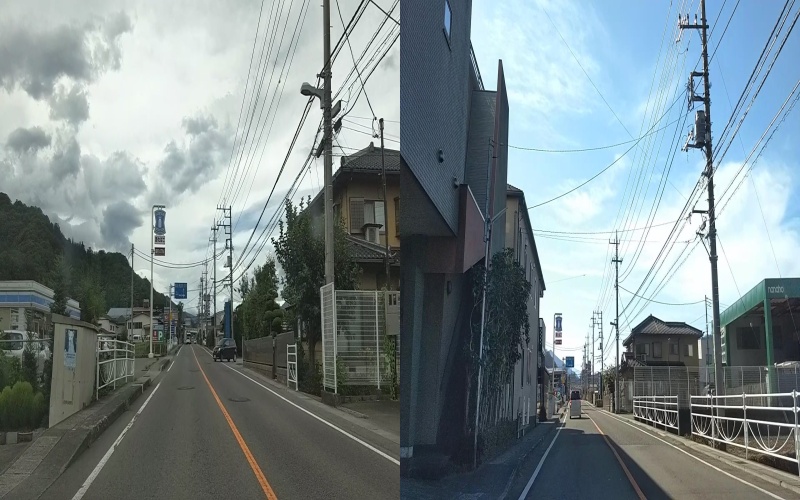}  
\caption{Tsuru, Category : Easy}  
\label{fig:Tsuru} 
\end{figure}

\begin{figure}[tb] 
\centering  
\includegraphics[width=0.9\linewidth]{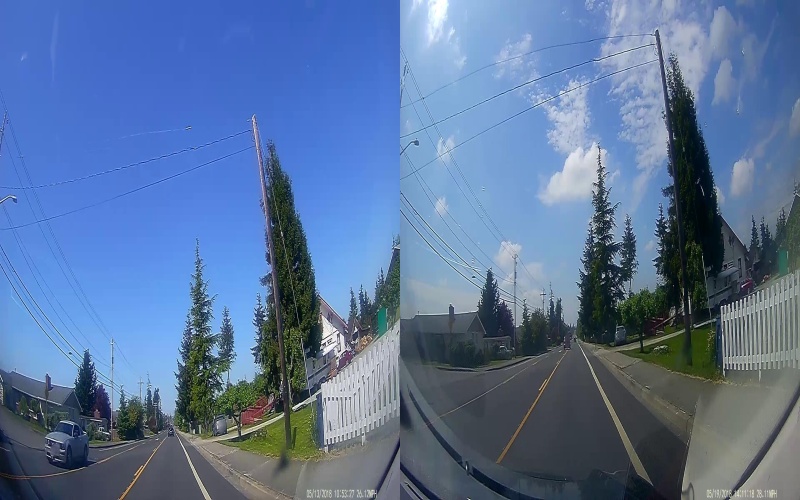}  
\caption{Washington, Category : Medium}  
\label{fig:Washington} 
\end{figure}

\begin{figure}[tb] 
\centering  
\includegraphics[width=0.9\linewidth]{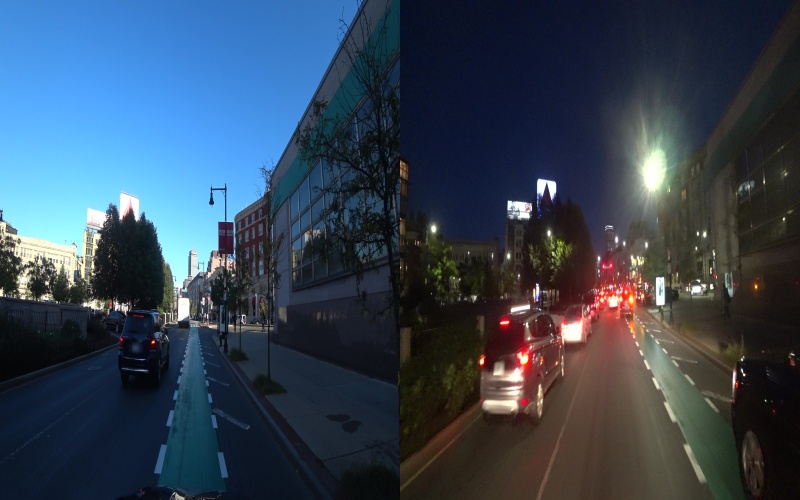}  
\caption{Boston1, Category : Hard}  
\label{fig:Boston1} 
\end{figure}

\begin{figure}[tb] 
\centering  
\includegraphics[width=0.9\linewidth]{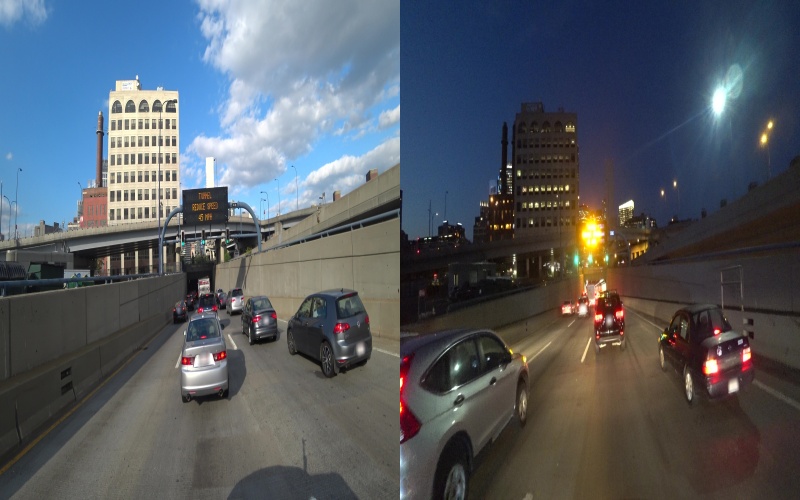}  
\caption{Boston2, Category : Medium}  
\label{fig:Boston2} 
\end{figure}

\begin{figure}[tb] 
\centering  
\includegraphics[width=0.9\linewidth]{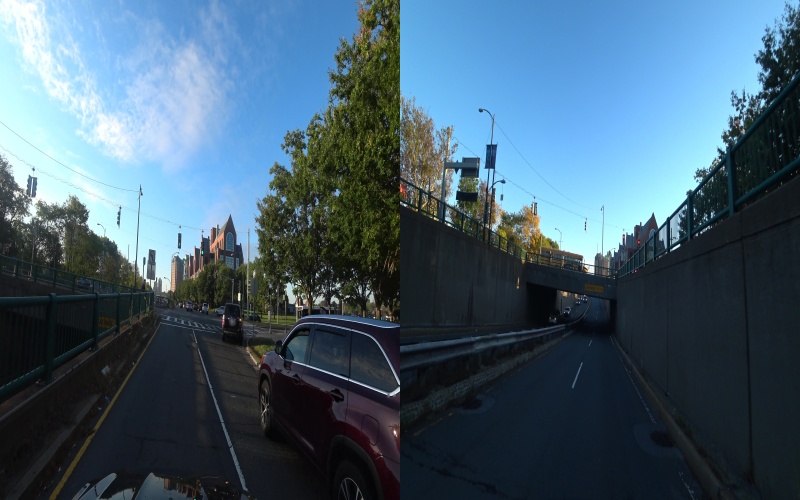}  
\caption{Boston3, Category : Medium}  
\label{fig:Boston3} 
\end{figure}

\begin{figure}[tb] 
\centering  
\includegraphics[width=0.9\linewidth]{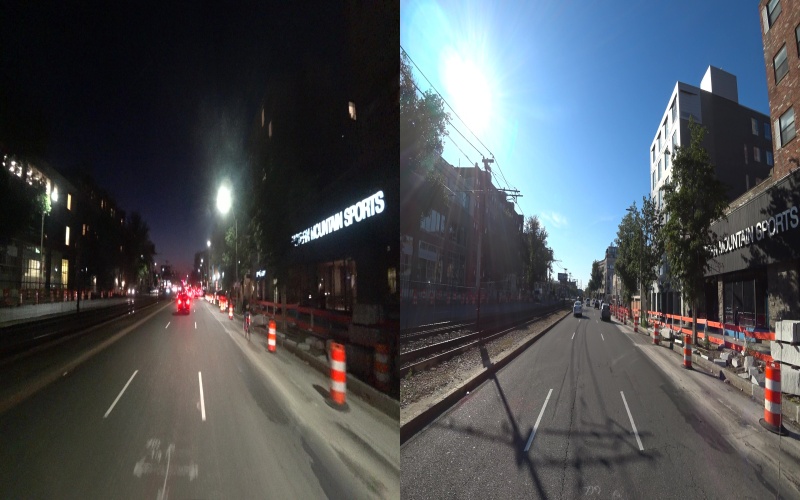}  
\caption{Boston4, Category : Medium}  
\label{fig:Boston4} 
\end{figure}

\begin{figure}[tb] 
\centering  
\includegraphics[width=0.9\linewidth]{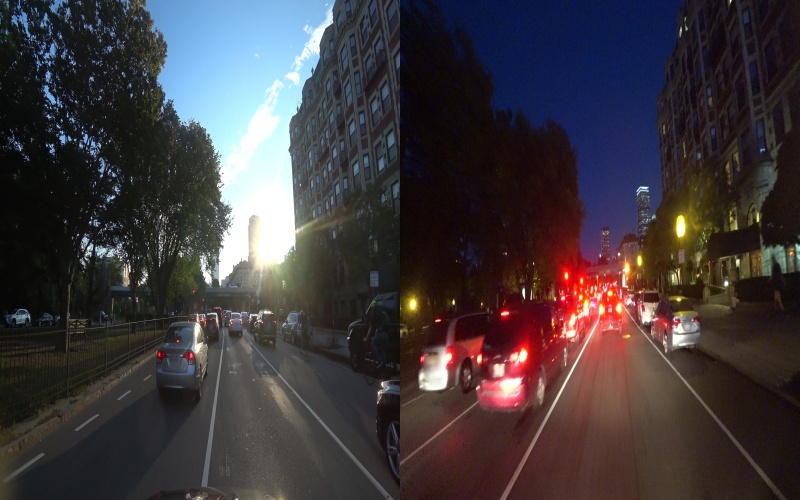}  
\caption{Boston5, Category : Medium}  
\label{fig:Boston5} 
\end{figure}

\begin{figure}[tb] 
\centering  
\includegraphics[width=0.9\linewidth]{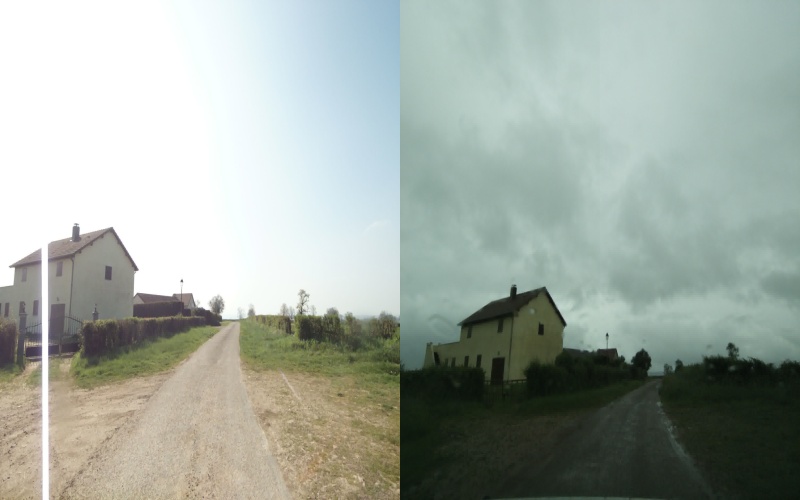}  
\caption{Burgundy2, Category : Medium}  
\label{fig:Burgundy2} 
\end{figure}

\begin{figure}[tb] 
\centering  
\includegraphics[width=0.9\linewidth]{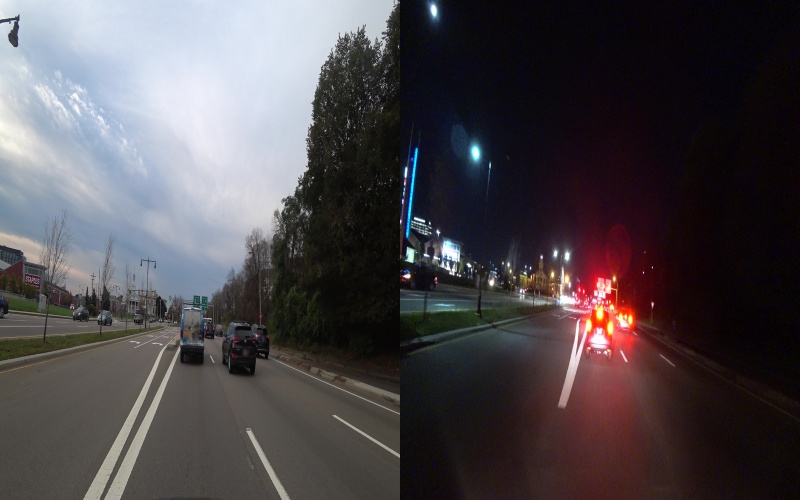}  
\caption{Massachusetts2, Category : Hard}  
\label{fig:Massachusetts2} 
\end{figure}

\begin{figure}[tb] 
\centering  
\includegraphics[width=0.9\linewidth]{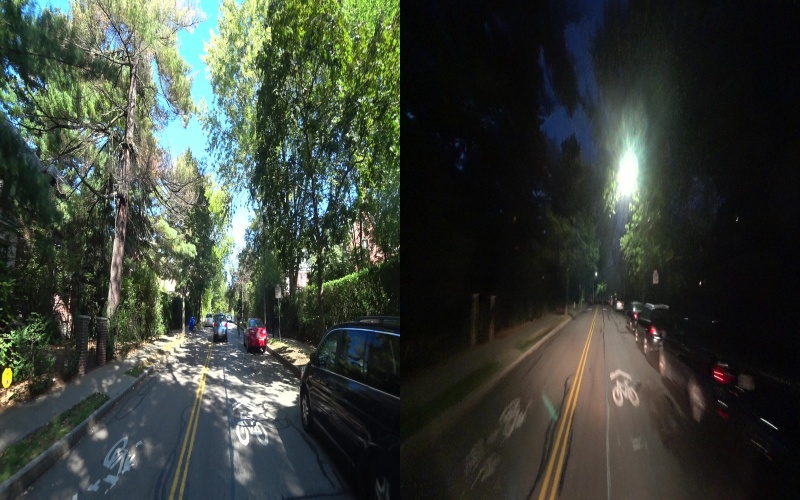}  
\caption{Massachusetts3, Category : Medium}  
\label{fig:Massachusetts3} 
\end{figure}

\begin{figure}[tb] 
\centering  
\includegraphics[width=0.9\linewidth]{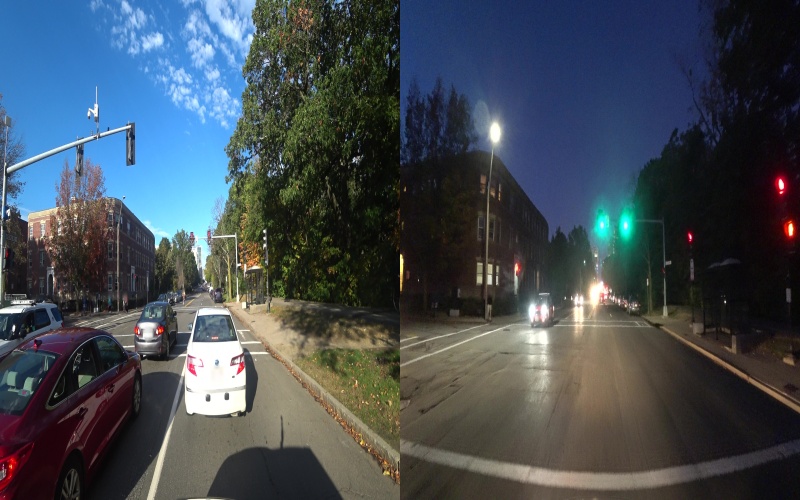}  
\caption{Massachusetts4, Category : Medium}  
\label{fig:Massachusetts4} 
\end{figure}

\begin{figure}[tb] 
\centering  
\includegraphics[width=0.9\linewidth]{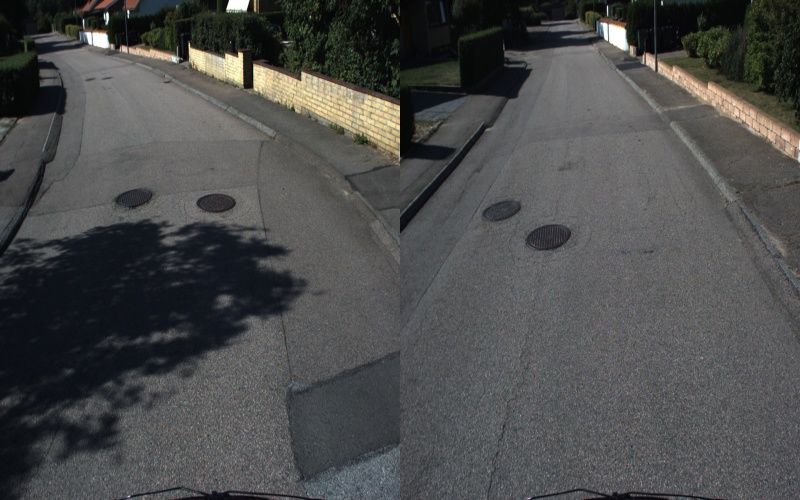}  
\caption{Skåne, Category : Medium}  
\label{fig:Skåne} 
\end{figure}

\begin{figure}[tb] 
\centering  
\includegraphics[width=0.9\linewidth]{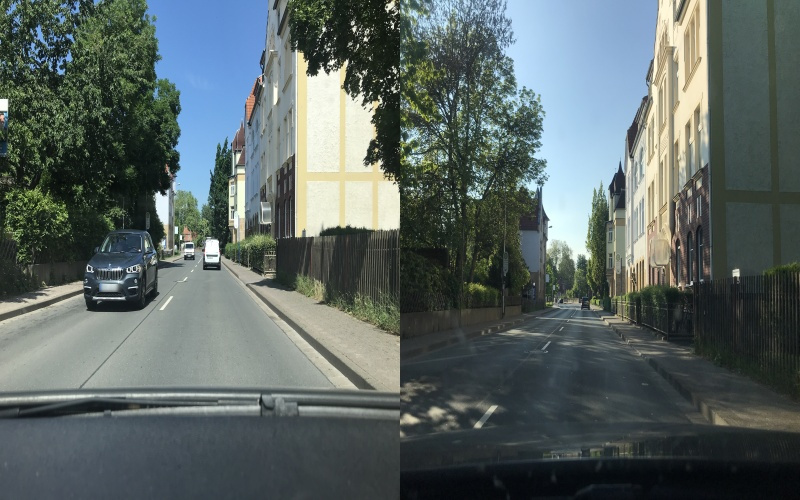}  
\caption{Thuringia, Category : Easy}  
\label{fig:Thuringia} 
\end{figure}

\begin{figure}[tb] 
\centering  
\includegraphics[width=0.9\linewidth]{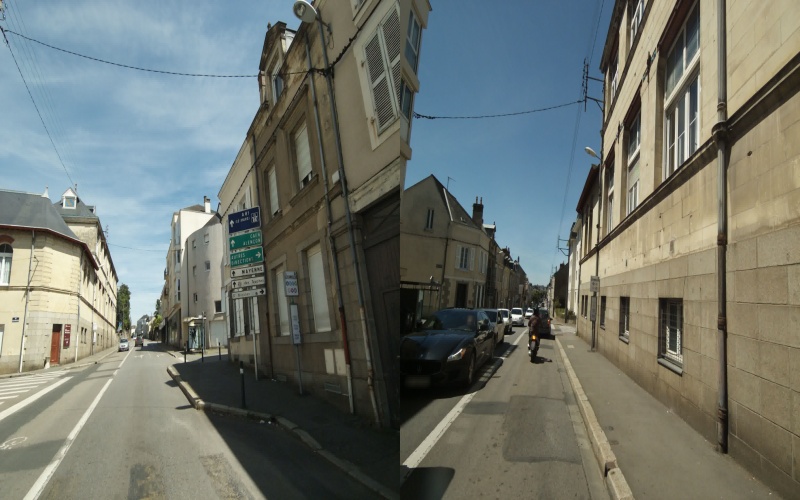}  
\caption{Angers1, Category : Difficult}  
\label{fig:Angers1} 
\end{figure}

\begin{figure}[tb] 
\centering  
\includegraphics[width=0.9\linewidth]{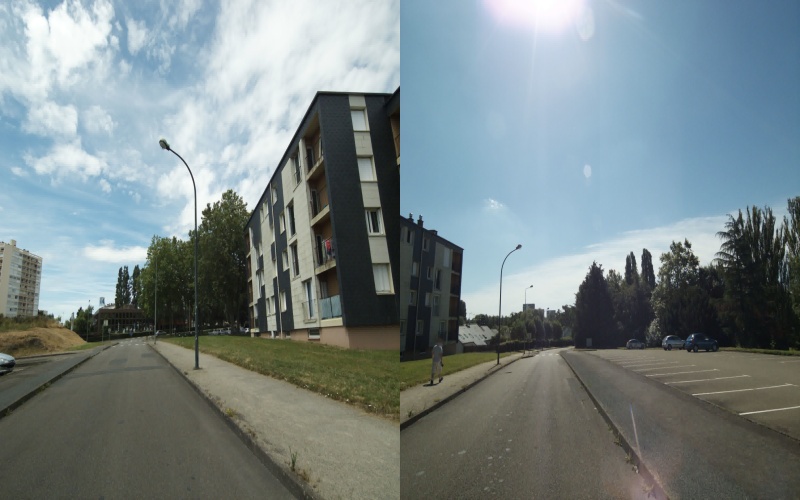}  
\caption{Angers2, Category : Difficult}  
\label{fig:Angers2} 
\end{figure}

\begin{figure}[tb] 
\centering  
\includegraphics[width=0.9\linewidth]{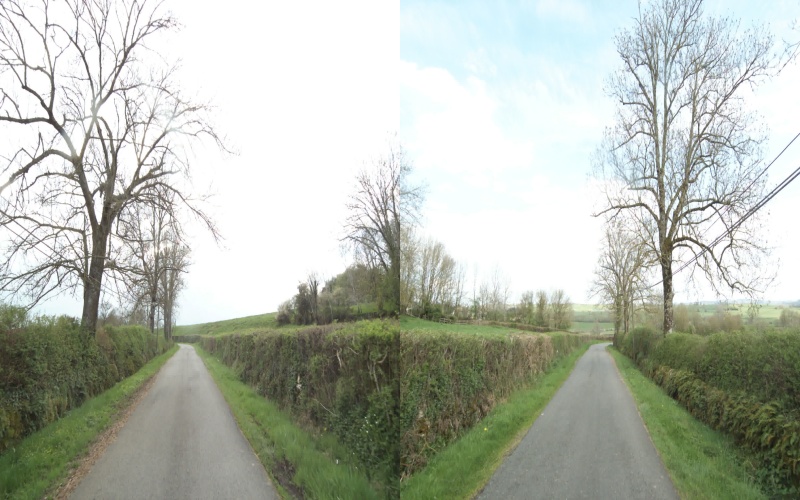}  
\caption{Besançon2, Category : Difficult}  
\label{fig:Besançon2} 
\end{figure}

\begin{figure}[tb] 
\centering  
\includegraphics[width=0.9\linewidth]{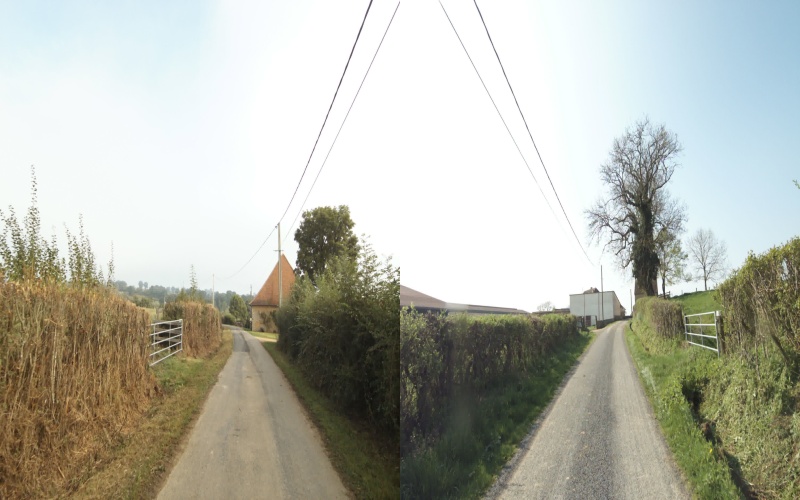}  
\caption{Besançon3, Category : Difficult}  
\label{fig:Besançon3} 
\end{figure}

\begin{figure}[tb] 
\centering  
\includegraphics[width=0.9\linewidth]{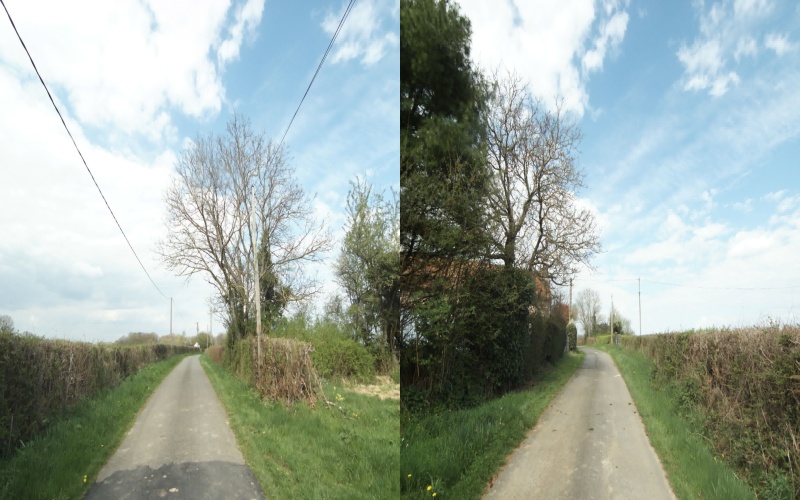}  
\caption{Besançon4, Category : Difficult}  
\label{fig:Besançon4} 
\end{figure}

\begin{figure}[tb] 
\centering  
\includegraphics[width=0.9\linewidth]{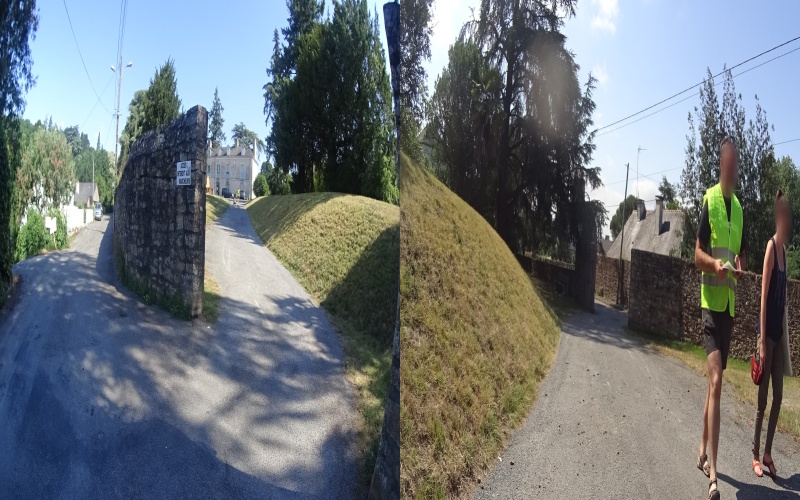}  
\caption{Brittany, Category : Difficult}  
\label{fig:Brittany} 
\end{figure}

\begin{figure}[tb] 
\centering  
\includegraphics[width=0.9\linewidth]{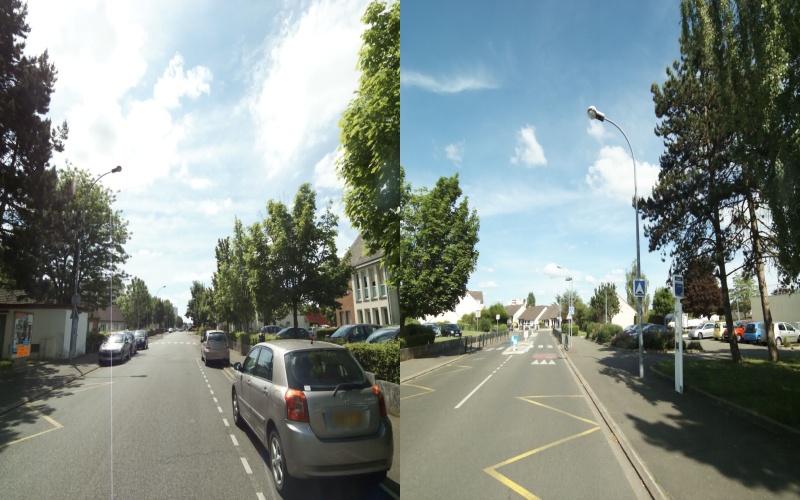}  
\caption{Brourges, Category : Difficult}  
\label{fig:Brourges} 
\end{figure}

\begin{figure}[tb] 
\centering  
\includegraphics[width=0.9\linewidth]{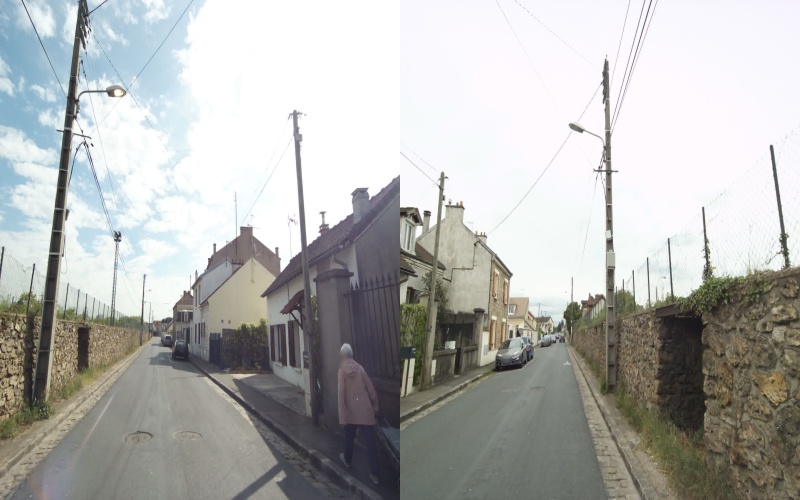}  
\caption{Ile-de-France, Category : Difficult}  
\label{fig:Ile-de-France} 
\end{figure}

\begin{figure}[tb] 
\centering  
\includegraphics[width=0.9\linewidth]{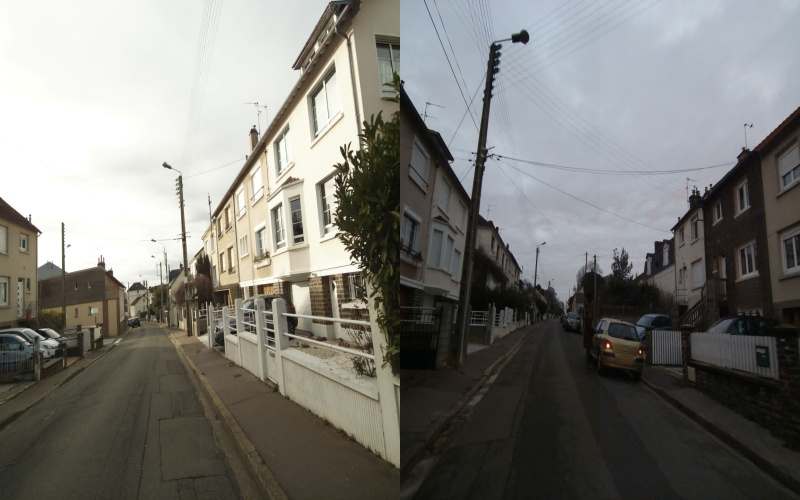}  
\caption{Le-Mans, Category : Difficult}  
\label{fig:Le-Mans} 
\end{figure}

\begin{figure}[tb] 
\centering  
\includegraphics[width=0.9\linewidth]{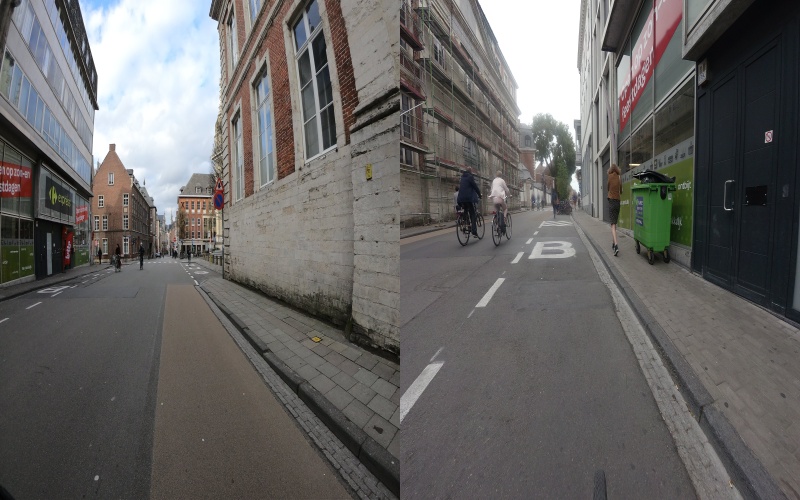}  
\caption{Leuven, Category : Difficult}  
\label{fig:Leuven} 
\end{figure}

\begin{figure}[tb] 
\centering  
\includegraphics[width=0.9\linewidth]{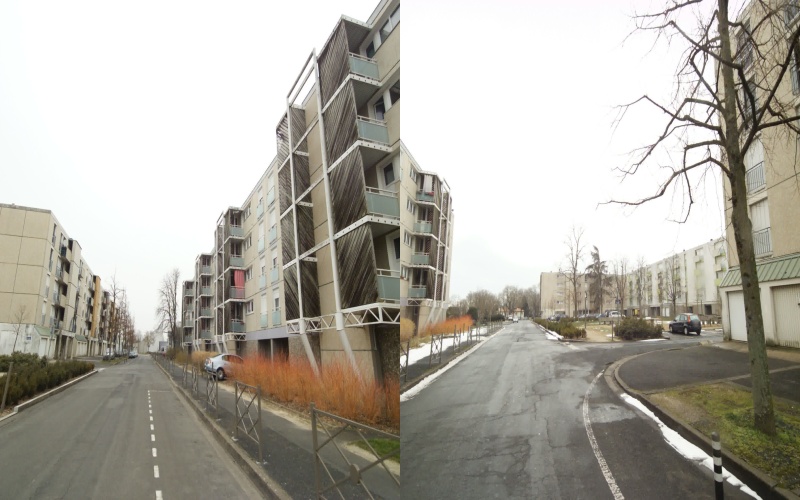}  
\caption{Nouvelle-Aquitaine1, Category : Difficult}  
\label{fig:Nouvelle-Aquitaine1} 
\end{figure}

\begin{figure}[tb] 
\centering  
\includegraphics[width=0.9\linewidth]{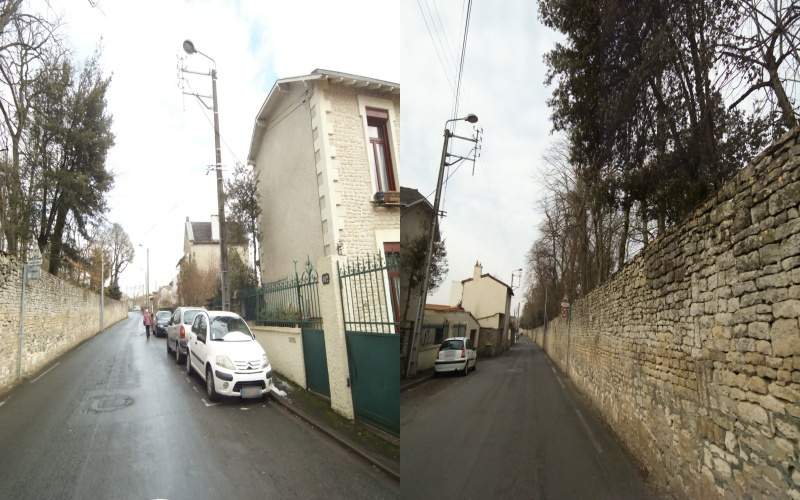}  
\caption{Nouvelle-Aquitaine2, Category : Difficult}  
\label{fig:Nouvelle-Aquitaine2} 
\end{figure}

\begin{figure}[tb] 
\centering  
\includegraphics[width=0.9\linewidth]{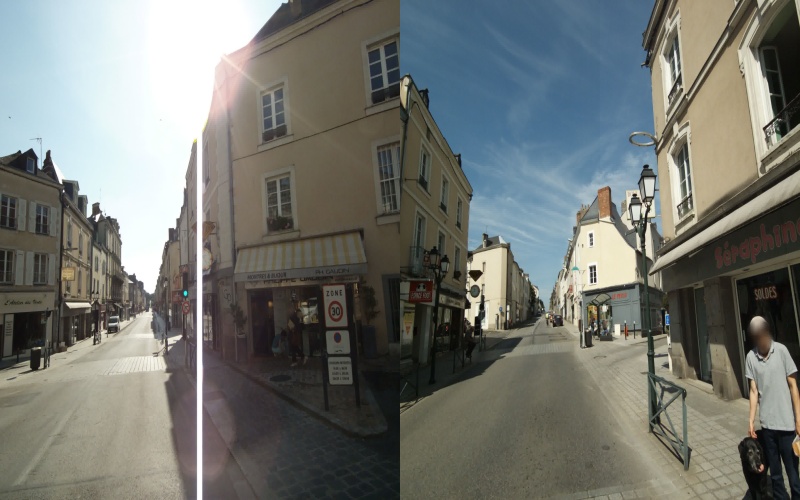}  
\caption{Orleans1, Category : Difficult}  
\label{fig:Orleans1} 
\end{figure}

\begin{figure}[tb] 
\centering  
\includegraphics[width=0.9\linewidth]{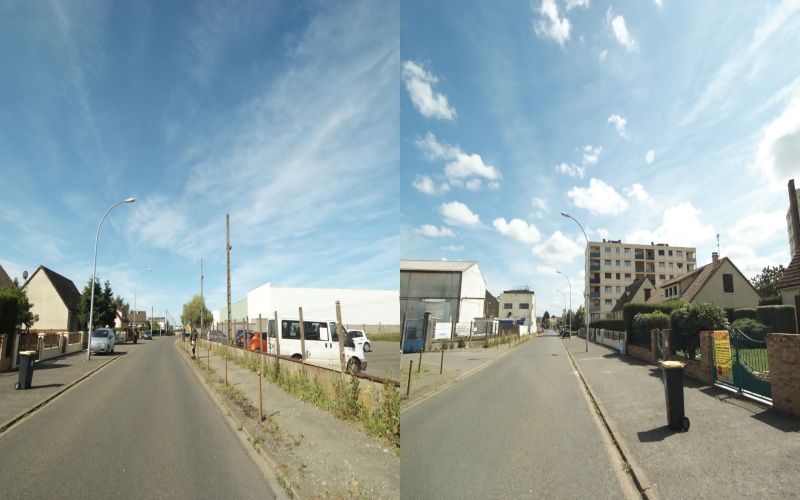}  
\caption{Orleans2, Category : Difficult}  
\label{fig:Orleans2} 
\end{figure}

\begin{figure}[tb] 
\centering  
\includegraphics[width=0.9\linewidth]{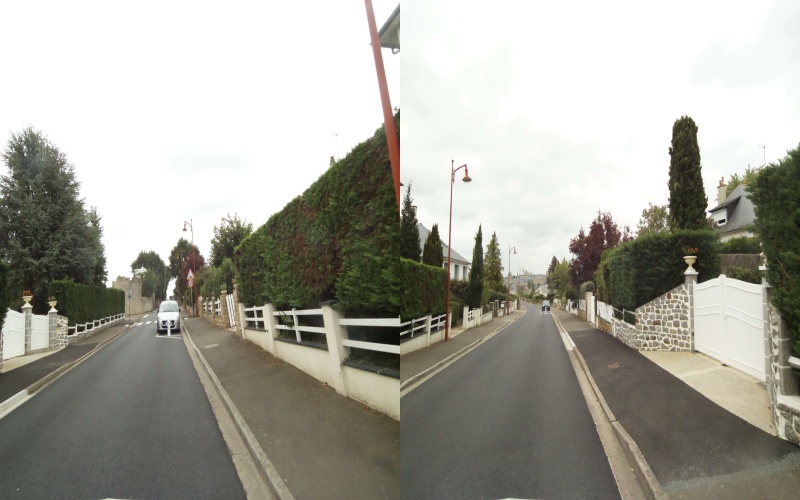}  
\caption{Pays de la Loire, Category : Difficult}  
\label{fig:Pays de la Loire} 
\end{figure}

In order to provide an overview over the type of scenes and conditions contained in our benchmark, Figures~\ref{fig:Bayern} to~\ref{fig:Pays de la Loire} show example images from all of our datasets and include the category (Easy, Medium, Hard) of the dataset. 
In each pair, the left image comes from the set of reference / training images, while the right comes from the set of query / test images. 
Note that for illustration purposes, the images have been resized to the same size. 
The aspect ratio in the figures thus differs from the aspect ratios of the images contained in the benchmark.



{\small
\bibliographystyle{ieee_fullname}
\bibliography{egbib}

\begin{thebibliography}{10}\itemsep=-1pt

\bibitem{OpenSfM}
{OpenSfM}.
\newblock \url{https://github.com/mapillary/OpenSfM}.

\bibitem{Waymo2019Open}
Waymo open dataset: An autonomous driving dataset.
\newblock \url{https://waymo.com/open/}, 2019.

\bibitem{ceres-solver}
Sameer Agarwal, Keir Mierle, and Others.
\newblock Ceres solver.
\newblock \url{http://ceres-solver.org}.

\bibitem{Arandjelovic2016CVPR}
Relja Arandjelovi\'c, Petr Gronat, Akihiko Torii, Tomas Pajdla, and Josef
  Sivic.
\newblock {NetVLAD}: {CNN} architecture for weakly supervised place
  recognition.
\newblock In {\em CVPR}, 2016.

\bibitem{Badino_IV11}
Hernan Badino, Daniel Huber, and Takeo Kanade.
\newblock Visual topometric localization.
\newblock In {\em Intelligent Vehicles Symposium (IV)}, 2011.

\bibitem{Brachmann2021ICCV}
Eric Brachmann, Martin Humenberger, Carsten Rother, and Torsten Sattler.
\newblock {On the Limits of Pseudo Ground Truth in Visual Camera
  Re-Localization}.
\newblock In {\em International Conference on Computer Vision (ICCV)}, 2021.

\bibitem{Brachmann2017CVPR}
Eric Brachmann, Alexander Krull, Sebastian Nowozin, Jamie Shotton, Frank
  Michel, Stefan Gumhold, and Carsten Rother.
\newblock {DSAC - Differentiable RANSAC for Camera Localization}.
\newblock In {\em CVPR}, 2017.

\bibitem{Brachmann2016CVPR}
Eric Brachmann, Frank Michel, Alexander Krull, Michael~Ying Yang, Stefan
  Gumhold, and Carsten Rother.
\newblock Uncertainty-driven 6d pose estimation of objects and scenes from a
  single rgb image.
\newblock In {\em CVPR}, 2016.

\bibitem{Brachmann2019ICCV}
Eric Brachmann and Carsten Rother.
\newblock {Expert Sample Consensus Applied to Camera Re-Localization}.
\newblock In {\em The IEEE International Conference on Computer Vision (ICCV)},
  2019.

\bibitem{Brachmann2020ARXIV}
Eric Brachmann and Carsten Rother.
\newblock {Visual Camera Re-Localization from RGB and RGB-D Images Using DSAC}.
\newblock {\em arXiv preprint arXiv:2002.12324}, 2020.

\bibitem{Brahmbhatt2018CVPR}
Samarth Brahmbhatt, Jinwei Gu, Kihwan Kim, James Hays, and Jan Kautz.
\newblock Geometry-aware learning of maps for camera localization.
\newblock In {\em The IEEE Conference on Computer Vision and Pattern
  Recognition (CVPR)}, 2018.

\bibitem{Caesar2019CoRR}
Holger Caesar, Varun Bankiti, Alex~H. Lang, Sourabh Vora, Venice~Erin Liong,
  Qiang Xu, Anush Krishnan, Yu Pan, Giancarlo Baldan, and Oscar Beijbom.
\newblock nuscenes: {A} multimodal dataset for autonomous driving.
\newblock {\em CoRR}, abs/1903.11027, 2019.

\bibitem{Camposeco2016ECCV}
Federico Camposeco, Torsten Sattler, and Marc Pollefeys.
\newblock {Minimal Solvers for Generalized Pose and Scale Estimation from Two
  Rays and One Point}.
\newblock In {\em ECCV}, 2016.

\bibitem{carlevaris2016university}
Nicholas Carlevaris-Bianco, Arash~K Ushani, and Ryan~M Eustice.
\newblock University of michigan north campus long-term vision and lidar
  dataset.
\newblock {\em The International Journal of Robotics Research},
  35(9):1023--1035, 2016.

\bibitem{Cavallari20193DV}
Tommaso {Cavallari}, Luca {Bertinetto}, Jishnu {Mukhoti}, Philip {Torr}, and
  Stuart {Golodetz}.
\newblock Let's take this online: Adapting scene coordinate regression network
  predictions for online rgb-d camera relocalisation.
\newblock In {\em 3DV}, 2019.

\bibitem{Cavallari2017CVPR}
Tommaso Cavallari, Stuart Golodetz, Nicholas~A. Lord, Julien Valentin, Luigi
  Di~Stefano, and Philip H.~S. Torr.
\newblock {On-The-Fly Adaptation of Regression Forests for Online Camera
  Relocalisation}.
\newblock In {\em CVPR}, 2017.

\bibitem{chen2011city}
David~M Chen, Georges Baatz, Kevin K{\"o}ser, Sam~S Tsai, Ramakrishna
  Vedantham, Timo Pylv{\"a}n{\"a}inen, Kimmo Roimela, Xin Chen, Jeff Bach, Marc
  Pollefeys, et~al.
\newblock City-scale landmark identification on mobile devices.
\newblock In {\em CVPR 2011}, pages 737--744. IEEE, 2011.

\bibitem{Chen2017ICRA}
Zetao Chen, Adam Jacobson, Niko S{\"{u}}nderhauf, Ben Upcroft, Lingqiao Liu,
  Chunhua Shen, Ian~D. Reid, and Michael Milford.
\newblock {Deep Learning Features at Scale for Visual Place Recognition}.
\newblock {\em ICRA}, 2017.

\bibitem{DeTone2018CVPRW}
D. {DeTone}, T. {Malisiewicz}, and A. {Rabinovich}.
\newblock {SuperPoint: Self-Supervised Interest Point Detection and
  Description}.
\newblock In {\em IEEE/CVF Conference on Computer Vision and Pattern
  Recognition Workshops (CVPRW)}, 2018.

\bibitem{doan2020visual}
Anh-Dzung Doan, Yasir Latif, Tat-Jun Chin, Yu Liu, Shin-Fang Ch’ng,
  Thanh-Toan Do, and Ian Reid.
\newblock Visual localization under appearance change: filtering approaches.
\newblock {\em Neural Computing and Applications}, pages 1--14, 2020.

\bibitem{doan2019scalable}
Anh-Dzung Doan, Yasir Latif, Tat-Jun Chin, Yu Liu, Thanh-Toan Do, and Ian Reid.
\newblock Scalable place recognition under appearance change for autonomous
  driving.
\newblock In {\em Proceedings of the IEEE/CVF International Conference on
  Computer Vision}, pages 9319--9328, 2019.

\bibitem{Dong2021CVPR}
Siyan Dong, Qingnan Fan, He Wang, Ji Shi, Li Yi, Thomas Funkhouser, Baoquan
  Chen, and Leonidas~J. Guibas.
\newblock Robust neural routing through space partitions for camera
  relocalization in dynamic indoor environments.
\newblock In {\em Proceedings of the IEEE/CVF Conference on Computer Vision and
  Pattern Recognition (CVPR)}, 2021.

\bibitem{dusmanu2019d2}
Mihai Dusmanu, Ignacio Rocco, Tomas Pajdla, Marc Pollefeys, Josef Sivic,
  Akihiko Torii, and Torsten Sattler.
\newblock D2-net: A trainable cnn for joint detection and description of local
  features.
\newblock {\em arXiv preprint arXiv:1905.03561}, 2019.

\bibitem{Fischler81CACM}
M. Fischler and R. Bolles.
\newblock {Random Sampling Consensus: A Paradigm for Model Fitting with
  Application to Image Analysis and Automated Cartography}.
\newblock {\em {Communications of the ACM}}, 24:381--395, 1981.

\bibitem{ForstnerWrobel}
Wolfgang F\"{o}rstner and Bernhard~P. Wrobel.
\newblock {\em {Photogrammetric Computer Vision}}.
\newblock Springer International Publishing, 2016.

\bibitem{Frahm10ECCV}
Jan-Michael Frahm, Pierre Fite-Georgel, David Gallup, Tim Johnson, Rahul
  RaguramC, hangchang Wu, Yi-Hung Jen, Enrique Dunn, Brian Clipp, Svetlana
  Lazebnik, and Marc Pollefeys.
\newblock {Building Rome on a Cloudless Day}.
\newblock In {\em {European Conference on Computer Vision}}, 2010.

\bibitem{garg2021your}
Sourav Garg, Tobias Fischer, and Michael Milford.
\newblock Where is your place, visual place recognition?
\newblock {\em arXiv preprint arXiv:2103.06443}, 2021.

\bibitem{garg2019semantic}
Sourav Garg, Niko Suenderhauf, and Michael Milford.
\newblock Semantic--geometric visual place recognition: a new perspective for
  reconciling opposing views.
\newblock {\em The International Journal of Robotics Research}, page
  0278364919839761, 2019.

\bibitem{garg2021semantics}
Sourav Garg, Niko S{\"u}nderhauf, Feras Dayoub, Douglas Morrison, Akansel
  Cosgun, Gustavo Carneiro, Qi Wu, Tat-Jun Chin, Ian Reid, Stephen Gould,
  et~al.
\newblock Semantics for robotic mapping, perception and interaction: A survey.
\newblock {\em arXiv preprint arXiv:2101.00443}, 2021.

\bibitem{ge-eccv-2020}
Yixiao Ge, Haibo Wang, Feng Zhu, Rui Zhao, and Hongsheng Li.
\newblock Self-supervising fine-grained region similarities for large-scale
  image localization.
\newblock In {\em Proceedings of the European Conference on Computer Vision
  (ECCV)}, 2020.

\bibitem{germain2019sparse}
Hugo Germain, Guillaume Bourmaud, and Vincent Lepetit.
\newblock Sparse-to-dense hypercolumn matching for long-term visual
  localization.
\newblock In {\em 2019 International Conference on 3D Vision (3DV)}, pages
  513--523. IEEE, 2019.

\bibitem{7scenes}
Ben Glocker, Shahram Izadi, Jamie Shotton, and Antonio Criminisi.
\newblock Real-time rgb-d camera relocalization.
\newblock In {\em International Symposium on Mixed and Augmented Reality
  (ISMAR)}. IEEE, October 2013.

\bibitem{Haralick94IJCV}
Bert~M. Haralick, Chung-Nan Lee, Karsten Ottenberg, and Michael N\"{o}lle.
\newblock Review and analysis of solutions of the three point perspective pose
  estimation problem.
\newblock {\em IJCV}, 13(3):331--356, 1994.

\bibitem{hartley2013rotation}
Richard Hartley, Jochen Trumpf, Yuchao Dai, and Hongdong Li.
\newblock Rotation averaging.
\newblock {\em International journal of computer vision}, 103(3):267--305,
  2013.

\bibitem{12scenes}
Caner Hazirbas, Sebastian~G. Soyer, Maximilian~C. Staab, Laura Leal-Taixé, and
  Daniel Cremers.
\newblock Deep depth from focus.
\newblock In {\em Asian Conference on Computer Vision (ACCV)}, December 2018.

\bibitem{Heinly2015CVPR}
Jared Heinly, Johannes~L. Sch\"{o}nberger, Enrique Dunn, and Jan-Michael Frahm.
\newblock Reconstructing the world* in six days.
\newblock In {\em CVPR}, 2015.

\bibitem{apolloscape}
Xinyu Huang, Xinjing Cheng, Qichuan Geng, Binbin Cao, Dingfu Zhou, Peng Wang,
  Yuanqing Lin, and Ruigang Yang.
\newblock The apolloscape dataset for autonomous driving.
\newblock {\em arXiv: 1803.06184}, 2018.

\bibitem{irschara2009structure}
Arnold Irschara, Christopher Zach, Jan-Michael Frahm, and Horst Bischof.
\newblock From structure-from-motion point clouds to fast location recognition.
\newblock In {\em 2009 IEEE Conference on Computer Vision and Pattern
  Recognition}, pages 2599--2606. IEEE, 2009.

\bibitem{Kendall2017CVPR}
Alex Kendall and Roberto Cipolla.
\newblock Geometric loss functions for camera pose regression with deep
  learning.
\newblock In {\em CVPR}, 2017.

\bibitem{Kendall2015ICCV}
Alex Kendall, Matthew Grimes, and Roberto Cipolla.
\newblock {PoseNet: {A} Convolutional Network for Real-Time 6-DOF Camera
  Relocalization}.
\newblock In {\em ICCV}, 2015.

\bibitem{lyft2019}
R. Kesten, M. Usman, J. Houston, T. Pandya, K. Nadhamuni, A. Ferreira, M. Yuan,
  B. Low, A. Jain, P. Ondruska, S. Omari, S. Shah, A. Kulkarni, A. Kazakova, C.
  Tao, L. Platinsky, W. Jiang, and V. Shet.
\newblock Lyft level 5 av dataset.
\newblock \url{https://level5.lyft.com/dataset/}, 2019.

\bibitem{Kukelova2016CVPR}
Zuzana Kukelova, Jan Heller, and Andrew Fitzgibbon.
\newblock {Efficient Intersection of Three Quadrics and Applications in
  Computer Vision}.
\newblock In {\em The IEEE Conference on Computer Vision and Pattern
  Recognition (CVPR)}, 2016.

\bibitem{Lebeda2012BMVC}
Karel Lebeda, Juan E.~Sala Matas, and Ondřej Chum.
\newblock {Fixing the Locally Optimized RANSAC}.
\newblock In {\em British Machine Vision Conference (BMVC)}, 2012.

\bibitem{Lee2015IJRR}
Gim~Hee Lee, Bo Li, Marc Pollefeys, and Friedrich Fraundorfer.
\newblock {Minimal solutions for the multi-camera pose estimation problem}.
\newblock {\em IJRR}, 34(7):837--848, 2015.

\bibitem{li2010location}
Yunpeng Li, Noah Snavely, and Daniel~P Huttenlocher.
\newblock Location recognition using prioritized feature matching.
\newblock In {\em European conference on computer vision}, pages 791--804.
  Springer, 2010.

\bibitem{Li2012ECCV}
Yunpeng Li, Noag Snavely, Dan~P. Huttenlocher, and Pascal Fua.
\newblock {Worldwide Pose Estimation Using 3D Point Clouds}.
\newblock In {\em ECCV}, 2012.

\bibitem{MPSD_2020_ECCV}
Manuel Lopez-Antequera, Pau Gargallo, Markus Hofinger, Samuel Rota~Bulò, Yubin
  Kuang, and Peter Kontschieder.
\newblock Mapillary planet-scale depth dataset.
\newblock In {\em Proceedings of the European Conference on Computer Vision
  (ECCV)}, 2020.

\bibitem{Lowe04IJCV}
David~G. Lowe.
\newblock {Distinctive Image Features from Scale-Invariant Keypoints}.
\newblock {\em {IJCV}}, 60(2), 2004.

\bibitem{lowry2016visual}
Stephanie Lowry, Niko S{\"u}nderhauf, Paul Newman, John~J Leonard, David Cox,
  Peter Corke, and Michael~J Milford.
\newblock Visual place recognition: A survey.
\newblock {\em IEEE Transactions on Robotics}, 32(1):1--19, 2015.

\bibitem{Maddern2017IJRR}
Will Maddern, Geoff Pascoe, Chris Linegar, and Paul Newman.
\newblock {1 Year, 1000km: The Oxford RobotCar Dataset}.
\newblock {\em IJRR}, 36(1):3--15, 2017.

\bibitem{milford2012seqslam}
Michael~J Milford and Gordon~F Wyeth.
\newblock {SeqSLAM: Visual route-based navigation for sunny summer days and
  stormy winter nights}.
\newblock In {\em ICRA}, 2012.

\bibitem{naseer2018robust}
Tayyab Naseer, Wolfram Burgard, and Cyrill Stachniss.
\newblock Robust visual localization across seasons.
\newblock {\em IEEE Transactions on Robotics}, 34(2):289--302, 2018.

\bibitem{Pless2003CVPR}
R. Pless.
\newblock {Using Many Cameras as One}.
\newblock In {\em CVPR}, 2003.

\bibitem{sarlin2019coarse}
Paul-Edouard Sarlin, Cesar Cadena, Roland Siegwart, and Marcin Dymczyk.
\newblock From coarse to fine: Robust hierarchical localization at large scale.
\newblock In {\em Proceedings of the IEEE Conference on Computer Vision and
  Pattern Recognition}, pages 12716--12725, 2019.

\bibitem{sarlin2020superglue}
Paul-Edouard Sarlin, Daniel DeTone, Tomasz Malisiewicz, and Andrew Rabinovich.
\newblock {SuperGlue}: Learning feature matching with graph neural networks.
\newblock In {\em CVPR}, 2020.

\bibitem{Sarlin2021CVPR}
Paul-Edouard Sarlin, Ajaykumar Unagar, Mans Larsson, Hugo Germain, Carl Toft,
  Viktor Larsson, Marc Pollefeys, Vincent Lepetit, Lars Hammarstrand, Fredrik
  Kahl, and Torsten Sattler.
\newblock Back to the feature: Learning robust camera localization from pixels
  to pose.
\newblock In {\em Proceedings of the IEEE/CVF Conference on Computer Vision and
  Pattern Recognition (CVPR)}, 2021.

\bibitem{Sattler2016CVPR}
Torsten Sattler, Michal Havlena, Konrad Schindler, and Marc Pollefeys.
\newblock {Large-Scale Location Recognition and the Geometric Burstiness
  Problem}.
\newblock In {\em CVPR}, 2016.

\bibitem{Sattler2011ICCV}
Torsten Sattler, Bastian Leibe, and Leif Kobbelt.
\newblock {Fast Image-Based Localization using Direct 2D-to-3D Matching}.
\newblock In {\em ICCV}, 2011.

\bibitem{Sattler2012ECCV}
Torsten Sattler, Bastian Leibe, and Leif Kobbelt.
\newblock {Improving Image-Based Localization by Active Correspondence Search}.
\newblock In {\em ECCV}, 2012.

\bibitem{Sattler2017PAMI}
T. Sattler, B. Leibe, and L. Kobbelt.
\newblock {Efficient \& Effective Prioritized Matching for Large-Scale
  Image-Based Localization}.
\newblock {\em PAMI}, 39(9):1744--1756, 2017.

\bibitem{Sattler2018CVPR}
Torsten Sattler, Will Maddern, Carl Toft, Akihiko Torii, Lars Hammarstrand,
  Erik Stenborg, Daniel Safari, Masatoshi Okutomi, Marc Pollefeys, Josef Sivic,
  Fredrik Kahl, and Tomas Pajdla.
\newblock {Benchmarking 6DOF Urban Visual Localization in Changing Conditions}.
\newblock In {\em CVPR}, 2017.

\bibitem{Sattler2017CVPR}
Torsten Sattler, Akihiko Torii, Josef Sivic, Marc Pollefeys, Hajime Taira,
  Masatoshi Okutomi, and Tomas Pajdla.
\newblock {Are Large-Scale 3D Models Really Necessary for Accurate Visual
  Localization?}
\newblock In {\em CVPR}, 2017.

\bibitem{Sattler2012BMVC}
Torsten Sattler, Tobias Weyand, Bastian Leibe, and Leif Kobbelt.
\newblock {Image Retrieval for Image-Based Localization Revisited}.
\newblock In {\em {British Machine Vision Conference}}, 2012.

\bibitem{Sattler2019CVPR}
Torsten Sattler, Qunjie Zhou, Marc Pollefeys, and Laura Leal-Taixe.
\newblock Understanding the limitations of cnn-based absolute camera pose
  regression.
\newblock In {\em The IEEE Conference on Computer Vision and Pattern
  Recognition (CVPR)}, June 2019.

\bibitem{Schoenberger2016CVPR}
Johannes~L. Sch\"{o}nberger and Jan-Michael Frahm.
\newblock {Structure-From-Motion Revisited}.
\newblock In {\em CVPR}, June 2016.

\bibitem{schonberger2018semantic}
Johannes~L Sch{\"o}nberger, Marc Pollefeys, Andreas Geiger, and Torsten
  Sattler.
\newblock Semantic visual localization.
\newblock In {\em Proceedings of the IEEE Conference on Computer Vision and
  Pattern Recognition}, pages 6896--6906, 2018.

\bibitem{se2002mobile}
Stephen Se, David Lowe, and Jim Little.
\newblock Mobile robot localization and mapping with uncertainty using
  scale-invariant visual landmarks.
\newblock {\em The international Journal of robotics Research}, 21(8):735--758,
  2002.

\bibitem{Shotton2013CVPR}
Jamie Shotton, Ben Glocker, Christopher Zach, Shahram Izadi, Antonio Criminisi,
  and Andrew Fitzgibbon.
\newblock {Scene Coordinate Regression Forests for Camera Relocalization in
  RGB-D Images}.
\newblock In {\em CVPR}, 2013.

\bibitem{Snavely-IJCV08}
Noah Snavely, Steven~M. Seitz, and Richard Szeliski.
\newblock Modeling the world from internet photo collections.
\newblock {\em IJCV}, 80(2):189--210, 2008.

\bibitem{sunderhaufwe}
Niko S{\"u}nderhauf, Peer Neubert, and Peter Protzel.
\newblock Are we there yet? challenging seqslam on a 3000 km journey across all
  four seasons.

\bibitem{Svarm2017PAMI}
Linus Sv\"{a}rm, Olof Enqvist, Fredrik Kahl, and Magnus Oskarsson.
\newblock {City-Scale Localization for Cameras with Known Vertical Direction}.
\newblock {\em PAMI}, 39(7):1455--1461, 2017.

\bibitem{Sweeney2014ECCV}
Chris Sweeney, Victor Fragoso, Tobias H{\"o}llerer, and Matthew Turk.
\newblock {gDLS: A Scalable Solution to the Generalized Pose and Scale
  Problem}.
\newblock In {\em ECCV}. Springer, 2014.

\bibitem{Taira2018CVPR}
Hajime Taira, Masatoshi Okutomi, Torsten Sattler, Mircea Cimpoi, Marc
  Pollefeys, Josef Sivic, Tomas Pajdla, and Akihiko Torii.
\newblock {InLoc}: Indoor visual localization with dense matching and view
  synthesis.
\newblock In {\em {CVPR}}, 2018.

\bibitem{toft2020single}
Carl Toft, Daniyar Turmukhambetov, Torsten Sattler, Fredrik Kahl, and Gabriel~J
  Brostow.
\newblock Single-image depth prediction makes feature matching easier.
\newblock In {\em European Conference on Computer Vision}, pages 473--492.
  Springer, 2020.

\bibitem{Torii2015CVPR}
Akihiko Torii, Relja Arandjelovi\'c, Josef Sivic, Masatoshi Okutomi, and Tomas
  Pajdla.
\newblock {24/7 Place Recognition by View Synthesis}.
\newblock In {\em CVPR}, 2015.

\bibitem{torii11}
Akihiko Torii, Josef Sivic, and Tomas Pajdla.
\newblock Visual localization by linear combination of image descriptors.
\newblock In {\em Proceedings of the 2nd IEEE Workshop on Mobile Vision, with
  ICCV}, 2011.

\bibitem{torii2013visual}
Akihiko Torii, Josef Sivic, Tomas Pajdla, and Masatoshi Okutomi.
\newblock Visual place recognition with repetitive structures.
\newblock In {\em Proceedings of the IEEE conference on computer vision and
  pattern recognition}, pages 883--890, 2013.

\bibitem{triggs1999bundle}
Bill Triggs, Philip~F McLauchlan, Richard~I Hartley, and Andrew~W Fitzgibbon.
\newblock Bundle adjustment—a modern synthesis.
\newblock In {\em International workshop on vision algorithms}, pages 298--372.
  Springer, 1999.

\bibitem{Valentin2016}
Julien Valentin, Angela Dai, Matthias Nie{\ss}ner, Pushmeet Kohli, Philip Torr,
  Shahram Izadi, and Cem Keskin.
\newblock {Learning to Navigate the Energy Landscape}.
\newblock In {\em 3DV}, 2016.

\bibitem{Ventura2014CVPR}
Jonathan Ventura, Clemens Arth, Gerhard Reitmayr, and Dieter Schmalstieg.
\newblock {A Minimal Solution to the Generalized Pose-and-Scale Problem}.
\newblock In {\em CVPR}, 2014.

\bibitem{vysotska2015lazy}
Olga Vysotska and Cyrill Stachniss.
\newblock Lazy data association for image sequences matching under substantial
  appearance changes.
\newblock {\em IEEE Robotics and Automation Letters}, 1(1):213--220, 2015.

\bibitem{Walch2017ICCV}
Florian Walch, Caner Hazirbas, Laura Leal{-}Taix{\'{e}}, Torsten Sattler,
  Sebastian Hilsenbeck, and Daniel Cremers.
\newblock {Image-Based Localization Using LSTMs for Structured Feature
  Correlation}.
\newblock In {\em ICCV}, 2017.

\bibitem{wald-eccv-2020}
Johanna Wald, Torsten Sattler, Stuart Golodetz, Tommaso Cavallari, and Federico
  Tombari.
\newblock Beyond controlled environments: {3D} camera re-localization in
  changing indoor scenes.
\newblock In {\em Proceedings of the European Conference on Computer Vision
  (ECCV)}, 2020.

\bibitem{warburg2020mapillary}
Frederik Warburg, Soren Hauberg, Manuel L{\'o}pez-Antequera, Pau Gargallo,
  Yubin Kuang, and Javier Civera.
\newblock Mapillary street-level sequences: A dataset for lifelong place
  recognition.
\newblock In {\em Proceedings of the IEEE/CVF Conference on Computer Vision and
  Pattern Recognition}, pages 2626--2635, 2020.

\bibitem{wijmans17rgbd}
Erik Wijmans and Yasutaka Furukawa.
\newblock Exploiting 2d floorplan for building-scale panorama rgbd alignment.
\newblock In {\em Computer Vision and Pattern Recognition, {CVPR}}, 2017.

\bibitem{Zamir10ECCV}
Amir~R. Zamir and Mubarak Shah.
\newblock {Accurate Image Localization Based on Google Maps Street View}.
\newblock In {\em ECCV}, 2010.

\bibitem{Zeisl2015ICCV}
Bernhard Zeisl, Torsten Sattler, and Marc Pollefeys.
\newblock Camera pose voting for large-scale image-based localization.
\newblock In {\em ICCV}, 2015.

\bibitem{Zhang06TDPVT}
Wei Zhang and Jana Kosecka.
\newblock {Image based Localization in Urban Environments}.
\newblock In {\em 3DPVT}, 2006.

\bibitem{Zhang2020IJCV}
Zichao Zhang, Torsten Sattler, and Davide Scaramuzza.
\newblock {Reference Pose Generation for Long-term Visual Localization via
  Learned Features and View Synthesis}.
\newblock {\em {International Journal of Computer Vision}}, pages 1--1, 2020.

\end{thebibliography}
}

\end{document}